\documentclass[10pt,twocolumn,letterpaper]{article}

\usepackage{cvpr}
\usepackage{times}
\usepackage{epsfig}
\usepackage{graphicx}
\usepackage{amsmath}
\usepackage{amssymb}
\newcommand{\argmin}{\mathop{\mathrm{argmin}}}   
\usepackage{tabu}
\usepackage{multirow}
\usepackage{comment}
\newcommand*{\our}{JiGen\@\xspace}

\usepackage{pifont}
\usepackage{gensymb}
\usepackage[export]{adjustbox}
\usepackage{verbatimbox}

\usepackage[breaklinks=true,bookmarks=false]{hyperref}

\cvprfinalcopy 

\def\cvprPaperID{****} 

\begin{document}

\title{Domain Generalization by Solving Jigsaw Puzzles}

\author{Fabio M. Carlucci$^{1}$\thanks{This work was done while at University of Rome Sapienza, Italy}
\hspace{0.8cm}
Antonio D'Innocente$^{2,3}$
\hspace{0.8cm}
Silvia Bucci$^{3}$\vspace{1mm}\\
Barbara Caputo$^{3,4}$
\hspace{0.8cm}
Tatiana Tommasi$^{4}$ \vspace{4mm}\\
$^{1}$Huawei, London
\hspace{0.5cm}
$^{2}$University of Rome Sapienza, Italy\vspace{1mm}\\
$^{3}$Italian Institute of Technology
\hspace{0.5cm}
$^{4}$Politecnico di Torino, Italy \vspace{2mm}\\
{\tt\small fabio.maria.carlucci@huawei.com \hspace{0.2cm} \{antonio.dinnocente, silvia.bucci\}@iit.it}\\
{\tt\small \{barbara.caputo, tatiana.tommasi\}@polito.it}
}

\maketitle
\thispagestyle{empty}
\pagestyle{empty}

\begin{abstract}

Human adaptability relies crucially on the ability to learn and merge knowledge both from supervised and unsupervised learning: the parents point out few important concepts, but then the children fill in the gaps on their own. This is particularly effective, because supervised learning can never be exhaustive and thus learning autonomously allows to discover invariances and regularities that help to generalize.
In this paper we propose to apply a similar approach to the task of object recognition across domains: our model learns the semantic labels in a supervised fashion, and broadens its understanding of the data by learning from self-supervised signals how to solve a jigsaw puzzle on the same images. This secondary task helps the network to learn the concepts of spatial correlation while acting as a regularizer for the classification task. Multiple experiments on the PACS, VLCS, Office-Home and digits datasets confirm our intuition and show that this simple method  outperforms previous domain generalization and adaptation solutions. An ablation study further illustrates the inner workings of our approach.
\end{abstract}

\section{Introduction}
In the current gold rush towards artificial intelligent systems it is becoming more and more evident 
that there is little intelligence without the ability to transfer knowledge and generalize across tasks, 
domains and categories \cite{csurka_book}. 
A large portion of computer vision research is dedicated to supervised methods that show remarkable results with 
convolutional neural networks in well defined settings, but still struggle when attempting these types of generalizations.  
Focusing on the ability to generalize across domains, the community has attacked this issue so far 
mainly by \emph{supervised learning} processes that search for semantic spaces able to capture basic data 
knowledge regardless of the specific appearance of input images.
Existing methods range from decoupling image style from the shared object content \cite{Bousmalis:DSN:NIPS16}, 
to pulling data of different domains together and imposing adversarial conditions \cite{Li_2018_CVPR,Li_2018_ECCV}, 
up to generating new samples to better cover the space spanned by any future target \cite{DG_ICLR18,Volpi_2018_NIPS}.
With the analogous aim of getting general purpose feature embeddings, an alternative research direction has
been recently pursued in the area of \emph{unsupervised learning}. The main techniques are based on the 
definition of tasks useful to learn visual invariances and regularities captured by 
spatial co-location of patches \cite{NorooziF16,Cruz2017,Noroozi_2018_CVPR}, counting primitives \cite{learningtocount}, 
image coloring \cite{zhang2016colorful}, video frame ordering \cite{misra2016unsupervised,videosiccv15}
and other self-supervised signals.

\begin{figure}
    \centering
\includegraphics[width=0.48\textwidth]{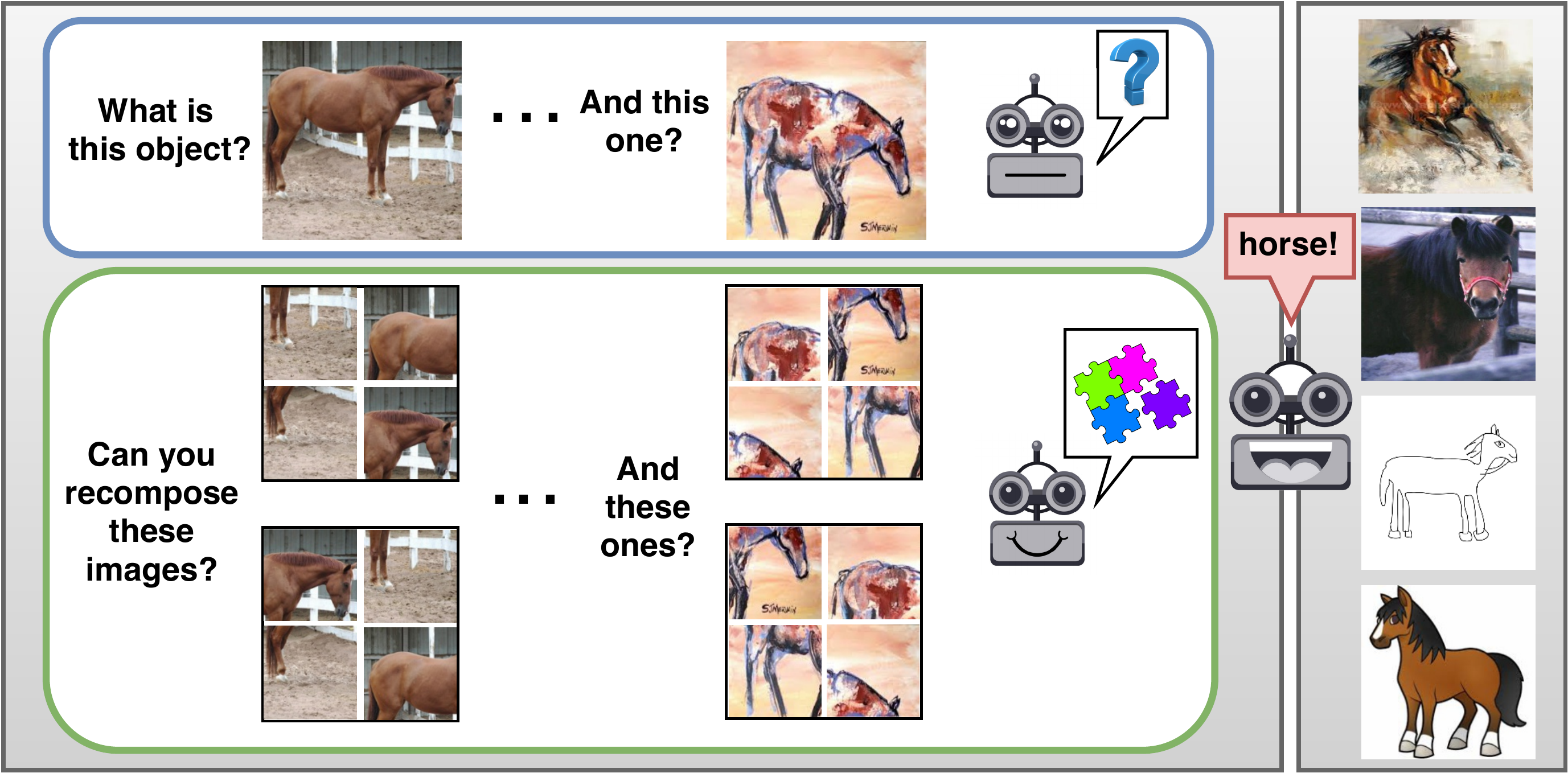}   
    \caption{Recognizing objects across visual domains is a challenging task that requires high generalization abilities.
    Other tasks, based on intrinsic self-supervisory image signals, allow to capture natural invariances and regularities
    that can help to bridge across large style gaps. With \our we learn jointly to classify objects and solve 
    jigsaw puzzles, showing that this supports generalization to new domains.}
    \label{fig:copertina}\vspace{-2mm}
\end{figure}

Since unlabeled data are largely available and by their very nature are less prone to bias (no labeling bias issue 
\cite{TorralbaEfros_bias}), they seem the perfect candidate to provide visual information independent from specific 
domain styles. 
Despite their large potential, the existing unsupervised approaches often come with tailored architectures
that need dedicated finetuning strategies to re-engineer the acquired knowledge and make it usable as input for a 
standard supervised training process \cite{Noroozi_2018_CVPR}.
Moreover, this knowledge is generally applied on real-world photos and has not been challenged before across
large domain gaps with images of other nature like paintings or sketches.

This clear separation between learning intrinsic regularities from images and robust classification across domains 
is in contrast with the visual learning strategies of biological systems, and in particular of the human visual system. 
Indeed, numerous studies highlight that infants and toddlers learn both to categorize objects and about regularities at 
the same time \cite{children_learning}. For instance, popular toys for infants teach to recognize different categories by fitting them 
into shape sorters; jigsaw puzzles of animals or vehicles to encourage learning of object parts' spatial relations 
are equally widespread among 12-18 months old. 
This type of joint learning is certainly a key ingredient in the ability of humans to reach sophisticated 
visual generalization abilities at an early age \cite{PLOS}. 

\begin{figure*}
    \centering
\includegraphics[width=0.9\textwidth]{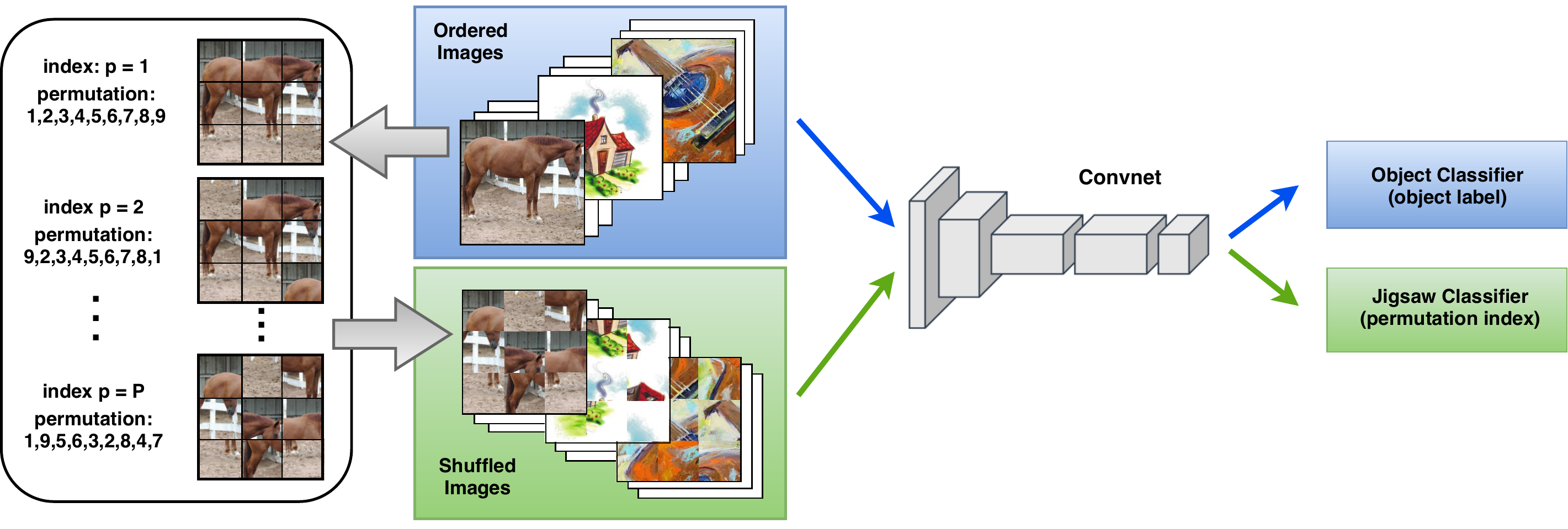}   
    \caption{Illustration of the proposed method \our. 
    We start from images of multiple domains and use a $3\times 3$ grid to decompose them in 9 patches
    which are then randomly shuffled and used to form images of the same dimension of the original ones.
    By using the maximal Hamming distance algorithm in \cite{NorooziF16} we define a set of $P$ patch permutations
    and assign an index to each of them. Both the original ordered and the shuffled images are fed to 
    a convolutional network that is optimized to satisfy two objectives: object classification on the 
    ordered images and jigsaw classification, meaning permutation index recognition, on the shuffled images.}
    \label{fig:jigen}\vspace{-3mm}
\end{figure*}

Inspired by this, we propose the first end-to-end architecture that learns simultaneously how to generalize across 
domains and about spatial co-location of image parts (Figure \ref{fig:copertina}, \ref{fig:jigen}). In this work we focus on the unsupervised task of 
recovering an original image from its shuffled parts, also known as \emph{solving jigsaw puzzles}. We show how this 
popular game can be re-purposed as a side objective to be optimized jointly with object classification over different 
source domains and improve generalization with a simple multi-task process \cite{Caruana:1997}.
We name our Jigsaw puzzle based Generalization method \emph{\our}.
Differently from previous approaches that deal with separate image patches and recombine their features towards the 
end of the learning process \cite{NorooziF16,Cruz2017,Noroozi_2018_CVPR}, we move the patch re-assembly at the 
image level and we formalize the jigsaw task as a classification problem over recomposed images with the same dimension 
of the original one. In this way object recognition and patch reordering can share the same network backbone and we
can seamlessly leverage over any convolutional learning structure as well as several pretrained models without the 
need of specific architectural changes.

We demonstrate that \our allows to better capture the shared knowledge among multiple sources
and acts as a regularization tool for a single source. In the case unlabeled samples of the target
data are available at training time, running the unsupervised jigsaw task on them 
contributes to the feature adaptation process and shows competing results with respect
to state of the art unsupervised domain adaptation methods.

\section{Related Work}
\paragraph{Solving Jigsaw Puzzles}
The task of recovering an original image from its shuffled parts is a basic pattern recognition problem that is commonly
identified with the jigsaw puzzle game. In the area of computer science and artificial intelligence it was first introduced by
\cite{FreemanG64}, which proposed a 9-piece puzzle solver based only on shape information and ignoring the image content.
Later, \cite{KosibaDBGK94} started to make use of both shape and appearance information. 
The problem has been mainly cast as predicting the permutations of a set of squared patches with all the challenges related 
to number and dimension of the patches,  their completeness (if all tiles are/aren't available) and homogeneity 
(presence/absence of extra tiles from other images). 
The application field for algorithms solving jigsaw puzzles is wide, from computer graphics and image editing 
\cite{patchtransformPAMI,SholomonDN14} to re-compose relics in archaeology \cite{Brown:2008:ASF,ECCV18_PaumardPT18}, 
from modeling in biology \cite{Marande415} to unsupervised learning of visual representations \cite{DoerschGE15,NorooziF16,Cruz2017}.
Existing assembly strategies can be broadly classified into two main categories: \emph{greedy} methods and \emph{global} methods. 
The first ones are based on sequential pairwise matches, while the second ones search for solutions that directly minimize a global 
compatibility measure over all the patches. 
Among the greedy methods, \cite{GallagherCVPR12} proposed a minimum spanning tree algorithm which progressively merges components 
while respecting the geometric consistent constraint. To eliminate matching outliers, \cite{loop_ECCV14} introduced loop
constraints among the patches. The problem can be also formulated as a classification task to predict the relative position 
of a patch with respect to another as in \cite{DoerschGE15}. Recently, \cite{ECCV18_PaumardPT18} expressed the patch reordering as 
the shortest path problem on a graph whose structure depends on the puzzle completeness and homogeneity.
The global methods consider all the patches together and use Markov Random Field formulations \cite{probabilistic_CVPR2010},
or exploit genetic algorithms \cite{SholomonDN14}. A condition on the consensus agreement among neighbors is used in 
\cite{CVPR16_SonMHC16}, while \cite{NorooziF16} focuses on a subset of possible permutations involving all the image tiles and solves a 
classification problem. The whole set of permutations is instead considered in \cite{Cruz2017} by approximating the permutation matrix 
and solving a bi-level optimization problem to recover the right ordering. 

Regardless of the specific approach and application, all the most recent deep-learning jigsaw-puzzle solvers tackle the problem by dealing 
with the separate tiles and then finding a way to recombine them. This implies designing tile-dedicated network architectures then 
followed by some specific process to transfer the collected knowledge in more standard settings that manage whole image samples.

\vspace{-4mm}\paragraph{Domain Generalization and Adaptation}
The goal of domain generalization (DG) is that of learning a system that can perform uniformly well across multiple 
data distributions. The main challenge is being able to distill the most useful and transferrable general knowledge 
from samples belonging to a limited number of population sources.
Several works have reduced the problem to the \emph{domain adaptation} (DA) setting where a fully labeled 
source dataset and an unlabeled set of examples from a different target domain are available \cite{csurka_book}.
In this case the provided target data is used to guide the source training procedure, that however
has to run again when changing the application target. To get closer to real world conditions, recent
work has started to focus on cases where the source data are drawn from multiple distributions \cite{mancini2018boosting,cocktail_CVPR18} 
and the target covers only a part of the source classes \cite{PADA_eccv18,LOAD_ICRA}.
For the more challenging DG setting with no target data available at training time, a large part of the 
previous literature presented \emph{model-based} strategies to neglect domain specific
signatures from multiple sources. They are both shallow and deep learning methods that 
build over multi-task learning \cite{ECCV12_Khosla}, low-rank network parameter decomposition \cite{hospedalesPACS} 
or domain specific aggregation layers \cite{Antonio_GCPR18}.  
Alternative solutions are based on source model weighting \cite{MassiRAL}, or on minimizing a validation 
measure on virtual tests defined from the available sources \cite{MLDG_AAA18}.
Other \emph{feature-level} approaches  search for a data representation able to capture information shared among multiple 
domains. This was formalized with the use of deep learning autoencoders in \cite{DGautoencoders, Li_2018_CVPR}, while \cite{doretto2017}
proposed to learn an embedding space where images of same classes but different sources are projected nearby. 
The recent work of \cite{Li_2018_ECCV} adversarially exploits class-specific domain classification modules to cover 
the cases where the covariate shift assumption does not hold and the sources have different class conditional distributions.
\emph{Data-level} methods propose to augment the source domain cardinality with the aim of covering a larger part of the data 
space and possibly get closer to the target. This solution was at first presented with the name of \emph{domain randomization} 
\cite{Tobin2017DomainRF} for samples from simulated environments whose variety was extended with random renderings. 
In \cite{DG_ICLR18} the augmentation is obtained with domain-guided perturbations of the original source instances.
Even when dealing with a single source domain, \cite{Volpi_2018_NIPS} showed that it is still possible to add adversarially 
perturbed samples by defining fictitious target distributions within a certain Wasserstein distance from the source. 

Our work stands in this DG framework, but proposes an orthogonal solution with respect to previous literature by
investigating the importance of jointly exploiting supervised and unsupervised inherent signals from the images.

\section{The \our Approach}
\label{sec:approach}
Starting from the samples of multiple source domains, we wish to learn a model that can perform well on any new 
target data population covering the same set of categories. 
Let us assume to observe $S$ domains, with the $i$-th domain containing  $N_i$ labeled instances 
$\{(x_j^{i},y_j^{i})\}_{j=1}^{N_i}$, where $x_j^{i}$ indicates the $j$-th image and $y_j^{i} \in \{1,\ldots,C\}$ 
is its class label. 
The first basic objective of \our is to 
minimize the loss  $\mathcal{L}_c(h(x| \theta_f, \theta_c),y)$ that measures the error between 
the true label $y$ and the label predicted by the deep model function $h$, parametrized by $\theta_f$ and $\theta_c$. 
These parameters define the feature embedding space and the final classifier, respectively for the convolutional
and fully connected parts of the network.
Together with this objective, we ask the network to satisfy a second condition related to solving jigsaw puzzles.
We start by decomposing the source images using a regular $n\times n$ grid of patches, which are then shuffled
and re-assigned to one of the $n^2$ grid positions. Out of the $n^2!$ possible permutations we select a set 
of $P$ elements by following the Hamming distance based algorithm in \cite{NorooziF16}, and we assign an index to each entry. 
In this way we define a second classification task on $K_i$ labeled instances $\{(z_k^{i},p_k^{i})\}_{k=1}^{K_i}$,
where $z_k^{i}$ indicates the recomposed samples and $p_k^i \in \{1,\ldots,P\}$ the related permutation index,
for which we need to minimize the jigsaw loss  $\mathcal{L}_p(h(z| \theta_f, \theta_p),p)$.
Here the deep model function $h$ has the same structure used for object classification and shares with that
the parameters  $\theta_f$. The final fully connected layer dedicated to permutation recognition
is parametrized by $\theta_p$. 
Overall we train the network to obtain the optimal model through \vspace{-5mm}
\begin{align}
    \argmin_{\theta_f, \theta_c, \theta_p} \sum_{i=1}^S  & \sum_{j=1}^{N_i}  \mathcal{L}_c(h(x^i_j | \theta_f, \theta_c),y_j^i) +  \nonumber \\ 
     & \sum_{k=1}^{K_i}\alpha\mathcal{L}_p(h(z^i_k | \theta_f, \theta_p),p_k^i) 
\end{align}
where both  $\mathcal{L}_c$  and $\mathcal{L}_p$ are standard cross-entropy losses. We underline that the
jigsaw loss is also calculated on the ordered images. Indeed, the correct patch sorting corresponds to one of 
the possible permutations and we always include it in the considered subset $P$. On the other way round, the 
classification loss is not influenced by the shuffled images, since this would make object recognition tougher. 
At test time we use only the object classifier to predict on the new target images.

\vspace{-4mm}\paragraph{Extension to Unsupervised Domain Adaptation}
Thanks to the unsupervised nature of the jigsaw puzzle task, we can always extend \our to the unlabeled samples
of target domain when available at training time. This allows us to exploit the jigsaw task for unsupervised domain adaptation. 
In this setting, for the target ordered images we minimize the classifier prediction uncertainty through the empirical entropy loss $\mathcal{L}_E (x^t)= \sum_{y\in \mathcal{Y}} h(x^t|\theta_f, \theta_c)log\{h(x^t|\theta_f, \theta_c)\}$, while for the shuffled target images we 
keep optimizing the jigsaw loss $\mathcal{L}_p(h(z^t | \theta_f, \theta_p),p^t)$.

\vspace{-4mm}\paragraph{Implementation Details}
Overall \our\footnote{Code available at https://github.com/fmcarlucci/JigenDG} has two parameters related to 
how we define the jigsaw task, and three related to the 
learning process. The first two are respectively the grid size $n \times n$ used to define the image patches 
and the cardinality of the patch permutation subset $P$. As we will detail in the following section, \our is 
robust to these values and for all our experiments we kept them fixed, using $3 \times 3$ patch grids and $P=30$.
The remaining parameters are the weights $\alpha$ of the jigsaw loss, and $\eta$ assigned to the entropy
loss when included in the optimization process for unsupervised domain adaptation. The final third 
parameter regulates the data input process: the shuffled images enter the network together with the original 
ordered ones, hence each image batch contains both of them.  We define a \emph{data bias} parameter $\beta$ to 
specify their relative ratio. For instance $\beta=0.6$ means that for each batch, $60\%$ of the
images are ordered, while the remaining $40\%$ are shuffled. These last three parameters were chosen by cross
validation on a $10\%$ subset of the source images for each experimental setting.

We designed the \our network making it able to leverage over many possible convolutional deep architectures.
Indeed it is sufficient to remove the existing last fully connected layer of a network and substitute
it with the new object and jigsaw classification layers.
\our is trained with SGD solver, $30$ epochs, batch size $128$,  learning 
rate set to $0.001$ and stepped down to $0.0001$ after $80\%$ of the training epochs. 
We used a simple data augmentation protocol by randomly cropping the images to retain between $80-100\%$ and 
randomly applied horizontal flipping. Following \cite{Noroozi_2018_CVPR} we randomly ($10\%$ probability) 
convert an image tile to grayscale.

\section{Experiments}

\paragraph{Datasets}
To evaluate the performance of \our when training over multiple sources we considered three domain generalization datasets. \textbf{PACS} \cite{hospedalesPACS} covers $7$ object categories and $4$ domains (Photo, Art Paintings, Cartoon and Sketches). We followed the experimental protocol in \cite{hospedalesPACS} and trained our model considering three domains as source datasets and the remaining one as target. \textbf{VLCS} \cite{TorralbaEfros_bias} aggregates images of $5$ object categories shared by the PASCAL VOC 2007, LabelMe, Caltech and Sun datasets which are considered as $4$ separated domains. We followed the standard protocol of \cite{DGautoencoders} dividing each domain  into a training set ($70\%$) and a test set ($30\%$) by random selection from the overall dataset. The \textbf{Office-Home} dataset \cite{venkateswara2017Deep} contains 65 categories of daily objects from 4 domains: Art, Clipart, Product and Real-World. In particular Product images are from vendor websites and show a white background, while Real-World represents object images collected with a regular camera. For this dataset we used the same experimental protocol of \cite{Antonio_GCPR18}. Note that Office-Home and PACS are related in terms of domain types and it is useful to consider
both as test-beds to check if  \our scales when the number of categories changes from 7 to 65. Instead VLCS offers different challenges because 
it combines object categories from Caltech with scene images of the other domains. 

To understand if solving jigsaw puzzles supports generalization even when dealing with a single source, we extended our analysis 
to \textbf{digit classification} as in \cite{Volpi_2018_NIPS}. We trained a model on 10k digit samples of the MNIST dataset 
\cite{lecun1998gradient} and evaluated on the respective test sets of MNIST-M~\cite{Ganin:DANN:JMLR16} and SVHN~\cite{netzer2011reading}. 
To work with comparable datasets, all the images were resized to $32\times32$ treated as RGB.

\begin{table}[tb]
\begin{center} \small
\begin{tabular}{@{}c@{~~~}c@{~~~}c@{~~~}c@{~~~}c@{~~~}c|@{~~~}c}
\hline
\multicolumn{2}{c}{\textbf{PACS}}  & \textbf{art\_paint.} & \textbf{cartoon} &  \textbf{sketches} & \textbf{photo} &   \textbf{Avg.}\\ \hline
\multicolumn{7}{@{}c@{}}{\textbf{CFN - Alexnet} }\\
\hline
\multicolumn{2}{@{}c@{}}{\footnotesize{J-CFN-Finetune}} & 47.23  & \textbf{62.18}  & \textbf{58.03}  & 70.18  & 59.41\\
\multicolumn{2}{@{}c@{}}{\footnotesize{J-CFN-Finetune++}} &  51.14  &  58.83  &  54.85  &  73.44  &  59.57\\
\multicolumn{2}{@{}c@{}}{\footnotesize{C-CFN-Deep All}} &  59.69  &  59.88  &  45.66  &  \underline{85.42}  &  62.66\\
\multicolumn{2}{@{}c@{}}{\footnotesize{C-CFN-\our}} &  \textbf{60.68}  &  60.55  &  55.66  &  82.68  &  \textbf{64.89}\\
\hline
\multicolumn{7}{c}{\textbf{Alexnet}}\\
\hline
\multirow{2}{*}{\cite{hospedalesPACS}}  & Deep All & 63.30 & 63.13 & 54.07 & 87.70 & 67.05\\
& TF & 62.86 & 66.97 & 57.51  & \textbf{89.50} & 69.21\\
\hline
\multirow{3}{*}{\cite{Li_2018_ECCV}} & Deep All & 57.55  & 67.04  & 58.52  & 77.98  & 65.27 \\
& DeepC &  62.30 & 69.58  & 64.45  &  80.72 &  69.26\\
& CIDDG &  62.70 & 69.73 & 64.45  &  78.65 &  68.88\\
\hline
\multirow{2}{*}{\cite{MLDG_AAA18}} & Deep All & 64.91 & 64.28 & 53.08 & 86.67 & 67.24\\
 & MLDG & 66.23 & 66.88 & 58.96 &  88.00& 70.01\\
\hline
\multirow{2}{*}{\cite{Antonio_GCPR18}} & Deep All  & 64.44 & \underline{72.07} & 58.07 &  87.50 & 70.52 \\
& D-SAM & 63.87 & 70.70 & 64.66 & 85.55 & 71.20\\
\hline
& Deep All & 66.68 & 69.41 & 60.02 & \underline{89.98}  & 71.52\\
& \textbf{\our} & \textbf{67.63} & \textbf{71.71} & \textbf{65.18} & 89.00 & \textbf{73.38}\\
\hline
\multicolumn{7}{c}{\textbf{Resnet-18}}\\
\hline
 \multirow{2}{*}{\cite{Antonio_GCPR18}} & Deep All & 77.87 & \underline{75.89} & 69.27 &  95.19 & 79.55\\
 & D-SAM & 77.33 & 72.43 & \textbf{77.83} & 95.30 & \textbf{80.72}\\
\hline
 & Deep All & 77.85  & 74.86  & 67.74  & 95.73  & 79.05 \\
 & \textbf{\our}    & \textbf{79.42}  & \textbf{75.25}  & 71.35  &  \textbf{96.03} &  80.51\\
\hline
\end{tabular}
\caption{Domain Generalization results on PACS. The results of \our are average over three repetitions of each run. 
Each column title indicates the name of the domain used as target. We use the bold font to highlight the best results of the generalization methods, while we underline a result when it is higher than all the others despite produced by the na\"ive Deep All baseline.
\emph{Top}: comparison with previous methods that use the jigsaw task as a pretext to learn transferable features using a context-free siamese-ennead network (CFN). \emph{Center} and \emph{Bottom}: comparison of \our with several domain generalization methods when using respectively Alexnet and Resnet-18 architectures. 
}
\label{table:resultsDG_PACS}
\end{center}\vspace{-4mm}
\end{table}

\begin{figure}[t]
\begin{tabular}{@{}c@{~~}c@{}}
\includegraphics[height=3cm]{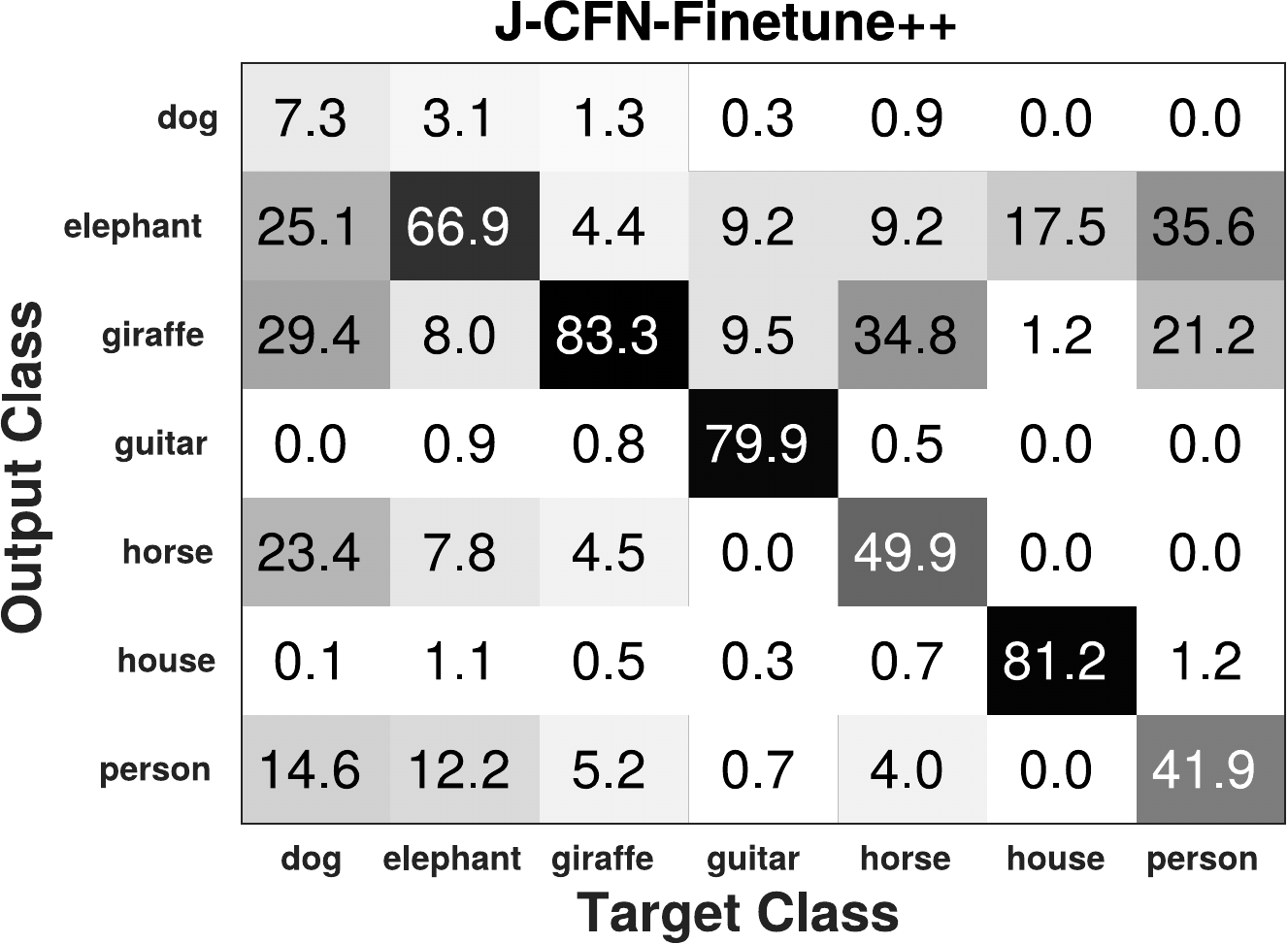} & 
\includegraphics[height=3cm]{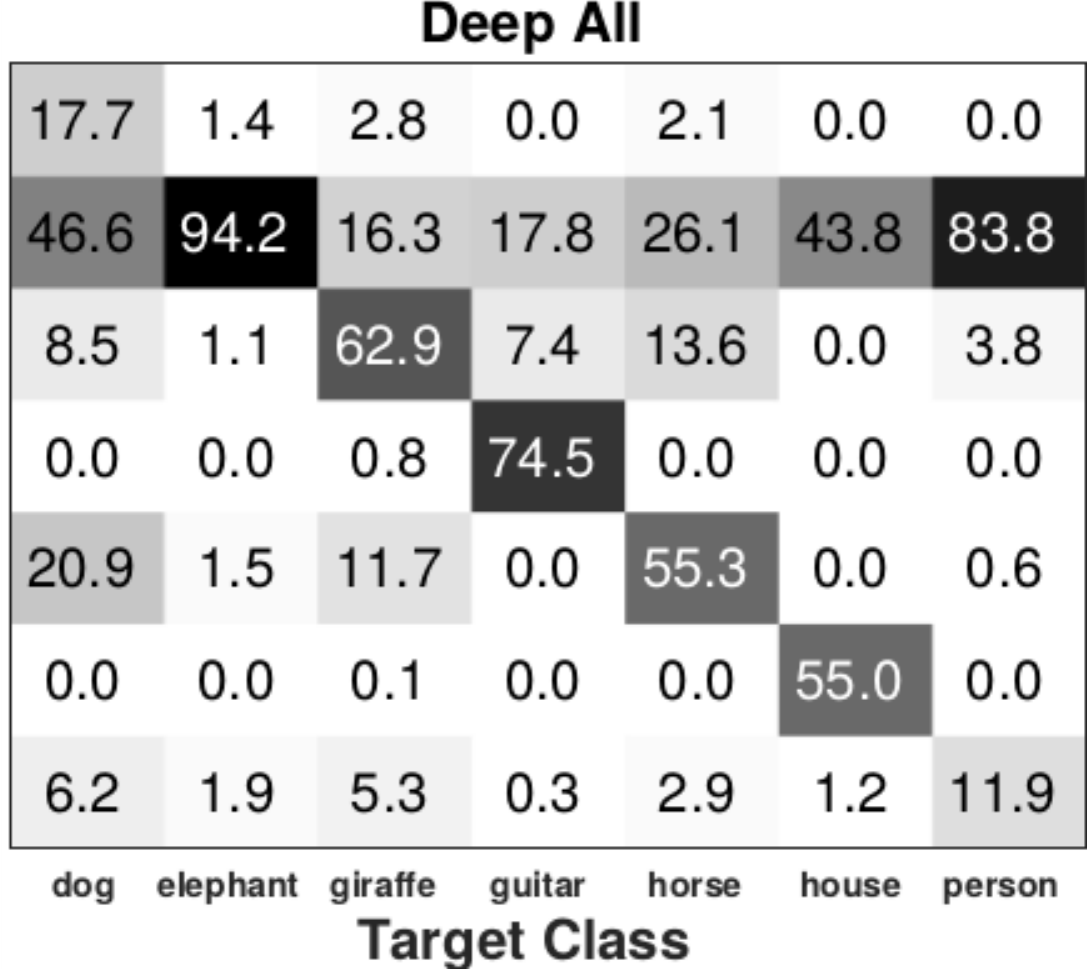} \\
\multicolumn{2}{c}{\includegraphics[height=3.1cm]{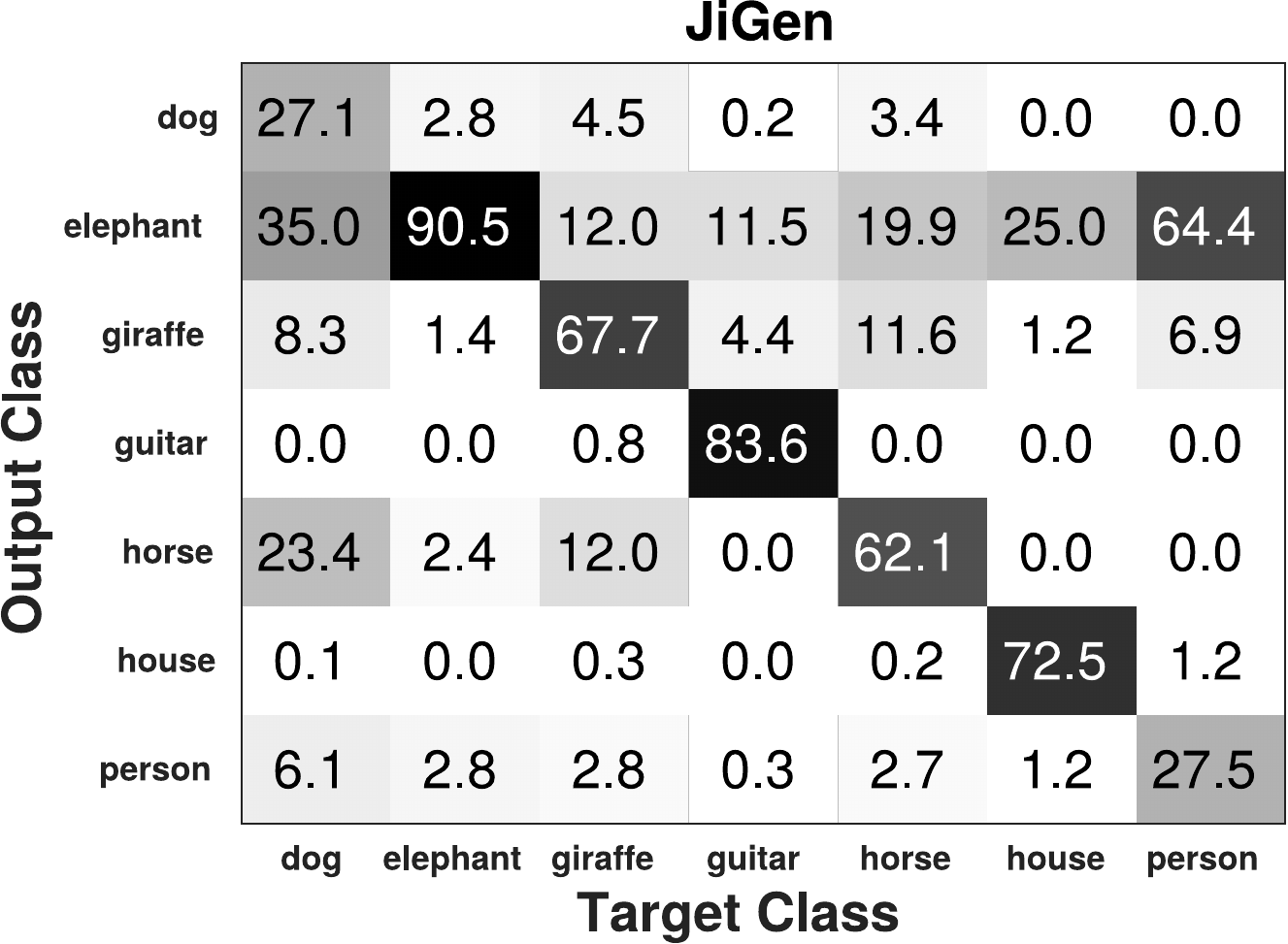}}
\end{tabular}
\caption{Confusion matrices on Alexnet-PACS DG setting, when sketches is used as target domain.}
\label{table:confmat}\vspace{-4mm}
\end{figure}

\vspace{-4mm}\paragraph{Patch-Based Convolutional Models for Jigsaw Puzzles}
We start our experimental analysis by evaluating the application of existing jigsaw related patch-based convolutional 
architectures and models to the domain generalization task.
We considered two recent works that proposed a jigsaw puzzle solver for 9 shuffled patches from images decomposed by a regular
$3\times3$ grid. Both \cite{NorooziF16} and \cite{Noroozi_2018_CVPR} use a Context-Free Network (CFN) with 9 siamese branches 
that extract features separately from each image patch and then recompose them before entering the final classification layer.
Specifically, each CFN branch is an Alexnet \cite{NIPS2012alexnet} up to the first fully connected layer ($fc6$) and all the branches 
share their weights. Finally, the branches' outputs are concatenated and given as input to the following 
fully connected layer ($fc7$). The jigsaw puzzle task is formalized as a classification problem on a subset
of patch permutations and, once the network is trained on a shuffled version of Imagenet \cite{imagenet}, the
learned weights can be used to initialize the $conv$ layers of a standard Alexnet while the rest of the network is trained 
from scratch for a new target task.  Indeed, according to the original works, the learned representation is able to capture
semantically relevant content from the images regardless of the object labels. 
We followed the instructions in \cite{NorooziF16} and started from the pretrained Jigsaw CFN (J-CFN) model provided by the authors
to run finetuning for classification on the PACS dataset with all the source domain samples aggregated together. 
In the top part of Table \ref{table:resultsDG_PACS} we indicate with \textbf{J-CFN-Finetune} the results of this experiment
using the jigsaw model proposed in \cite{NorooziF16}, while with \textbf{J-CFN-Finetune++} the results from the 
advanced model proposed in \cite{Noroozi_2018_CVPR}. 
In both cases the average classification accuracy on the domains is lower than what can be obtained with a standard Alexnet 
model pre-trained for object classification on Imagenet and finetuned on all the source data aggregated together.
We indicate this baseline approach with \emph{Deep All} and we can use as reference the corresponding values
in the following central part of Table \ref{table:resultsDG_PACS}. 
We can conclude that, despite its power as an unsupervised pretext task, completely disregarding 
the object labels when solving jigsaw puzzles induces a loss of semantic information that may be 
crucial for generalization across domains.

To demonstrate the potentialities of the CFN architecture, the authors of \cite{NorooziF16} used it also to train 
a supervised object Classification model on Imagenet (C-CFN) and demonstrated that it can produce results analogous to the
standard Alexnet. With the aim of further testing this network to understand if and how much its peculiar 
siamese-ennead structure can be useful to distill shared knowledge across domains, we considered it as the main 
convolutional backbone for \our.
Starting from the C-CFN model provided by the authors, we ran the obtained \textbf{C-CFN-\our} on 
PACS data, as well as its plain object classification version with the jigsaw loss disabled ($\alpha=0$) that we 
indicate as \textbf{C-CFN-Deep All}. From the obtained recognition accuracy we can state that combining the jigsaw puzzle with 
the classification task provides an average improvement in performance, which is the first result to confirm our intuition. 
However, C-CFN-Deep All is still lower than the reference results obtained with standard Alexnet.

For all the following experiments we consider the convolutional architecture of \our built with the same main 
structure of Alexnet or Resnet, using always the image as a whole (ordered or shuffled) instead of relying on 
separate patch-based network branches.
A detailed comparison of per-class results on the challenging sketches domain for J-CFN-Finetune++
and \our based on Alexnet reveals that for four out of seven categories, J-CFN-Finetune++  is actually doing a good job, 
better than Deep All. With \our we improve over Deep All for the same categories by solving jigsaw puzzles at image 
level and we keep the advantage of Deep All for the remaining categories.

\vspace{-4mm}\paragraph{Multi-Source Domain Generalization}
We compare the performance of \our against several recent domain generalization methods. 
\textbf{TF} is the low-rank parametrized network that was presented together with the dataset PACS in \cite{hospedalesPACS}. 
\textbf{CIDDG} is the conditional invariant deep domain generalization method presented in \cite{Li_2018_ECCV} that 
trains for image classification with two adversarial constraints: one that maximizes the overall domain 
confusion following \cite{Ganin:DANN:JMLR16} and a second one that does the same per-class. 
In the \textbf{DeepC} variant, only this second condition is enabled. 
\textbf{MLDG} \cite{MLDG_AAA18} is a meta-learning approach  that simulates 
train/test domain shift during training and exploit them to optimize the learning model. 
\textbf{CCSA} \cite{doretto2017}  learns an embedding subspace where mapped visual domains 
are semantically aligned and yet maximally separated. 
\textbf{MMD-AAE} \cite{Li_2018_CVPR} is a deep method based on adversarial autoencoders that learns an invariant feature representation 
by aligning the data distributions to an arbitrary prior through the Maximum  Mean Discrepancy (MMD).
\textbf{SLRC} \cite{Ding2017DeepDG} is based on a single domain invariant network and multiple domain specific ones and
it applies a low rank constraint among them. 
\textbf{D-SAM} \cite{Antonio_GCPR18} is a method based on the use of domain-specific aggregation modules combined to improve
model generalization: it provides the current sota results on PACS and Office-Home.
For each of these methods, the Deep All baseline indicates the performance of the corresponding network when all the introduced
domain adaptive conditions are disabled.

\begin{table}[tb]
\begin{center} \small
\begin{tabular}{@{}c@{~~~}c@{~~~}c@{~~~}c@{~~~}c@{~~~}c|c}
\hline
\multicolumn{2}{c}{\textbf{VLCS}}  & \textbf{Caltech} & \textbf{Labelme} &  \textbf{Pascal} & \textbf{Sun} &   \textbf{Avg.}\\ \hline
\multicolumn{7}{c}{\textbf{Alexnet}}\\
\hline
\multirow{3}{*}{\cite{Li_2018_ECCV}} & Deep All & 85.73	& 61.28	& 62.71	& 59.33	& 67.26  \\
& DeepC & 87.47	& 62.60	& 63.97	& 61.51	& 68.89 \\
& CIDDG & 88.83	& 63.06	& 64.38	& 62.10	& 69.59 \\
\hline
\multirow{2}{*}{\cite{doretto2017}} & Deep All & 86.10 & 55.60 & 59.10 & 54.60	& 63.85 \\
& CCSA & 92.30	& 62.10	& 67.10	& 59.10	& 70.15 \\
\hline
\multirow{2}{*}{\cite{Ding2017DeepDG}} & Deep All &  86.67 & 58.20 & 59.10 & 57.86 & 65.46 \\
& SLRC &  92.76	& 62.34	& 65.25	& 63.54	& 70.97 \\
\hline
\multirow{2}{*}{\cite{hospedalesPACS}} & Deep All & 93.40 & 62.11 & 68.41 & 64.16 & 72.02\\
 & TF & 93.63 & \textbf{63.49} & 69.99 & 61.32 & 72.11\\
\hline
 \cite{Li_2018_CVPR} & MMD-AAE & 94.40&	62.60&	67.70&	\textbf{64.40} &	72.28\\ 
 \hline
\multirow{2}{*}{\cite{Antonio_GCPR18}} & Deep All  & 94.95  & 57.45  & 66.06  & \underline{65.87}  &  71.08 \\
 & D-SAM & 91.75  & 56.95  & 58.59  & 60.84  & 67.03\\
\hline
 & Deep All & \underline{96.93} & 59.18 &  \underline{71.96} & 62.57 & 72.66\\
 & \textbf{\our} & \textbf{96.93} & 60.90 & \textbf{70.62} & 64.30 & \textbf{73.19}\\
\hline
\end{tabular}
\caption{Domain Generalization results on VLCS. For details about number of runs, meaning of columns and use of bold/underline fonts, see Table \ref{table:resultsDG_PACS}.}
\label{table:resultsDG_VLCS}\vspace{-4mm}
\end{center}
\end{table}

\begin{table}[tbp]
\begin{center}\small
\begin{tabular}{@{}c@{~~}c@{~~}c@{~~}c@{~~~}c@{~~~}c@{~}|@{~~}c@{~}}
\hline
 \multicolumn{2}{c}{\textbf{Office-Home}}  & \textbf{Art} & \textbf{Clipart} &  \textbf{Product} & \textbf{Real-World} &  \textbf{Avg.}\\ \hline
\multicolumn{7}{c}{\textbf{Resnet-18}}\\
\hline
 \multirow{2}{*}{\cite{Antonio_GCPR18}} & Deep All  & 55.59 & 42.42 & 70.34 & 70.86 & 59.81 \\
 & D-SAM & \textbf{58.03} & 44.37 & 69.22 & 71.45 & 60.77\\
\hline
 & Deep All & 52.15 & 45.86 &  70.86 & \underline{73.15} & 60.51\\
 & \textbf{\our} & 53.04 & \textbf{47.51} & \textbf{71.47} & 72.79 & \textbf{61.20}\\
\hline
\end{tabular}
\caption{Domain Generalization results on Office-Home. For details about number of runs, meaning of columns and use of bold/underline fonts, see Table \ref{table:resultsDG_PACS}.}
\label{table:resultsDG_officehome}
\end{center}\vspace{-8mm}
\end{table}

\begin{figure*}
    \centering
    \begin{tabular}{c@{~~~}c@{~~~}c@{~~~}c}
\includegraphics[height=3.3cm]{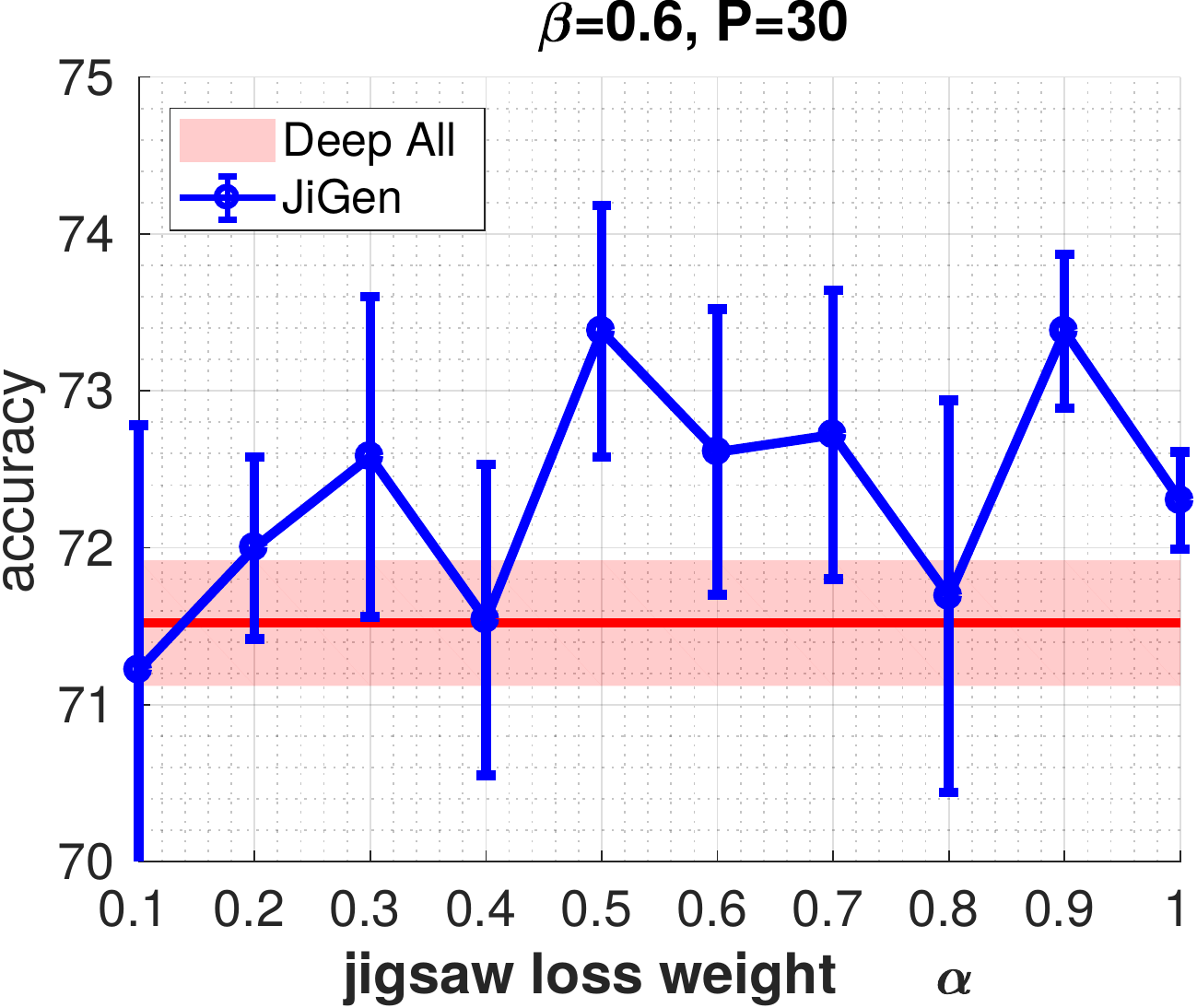}   &  \includegraphics[height=3.3cm]{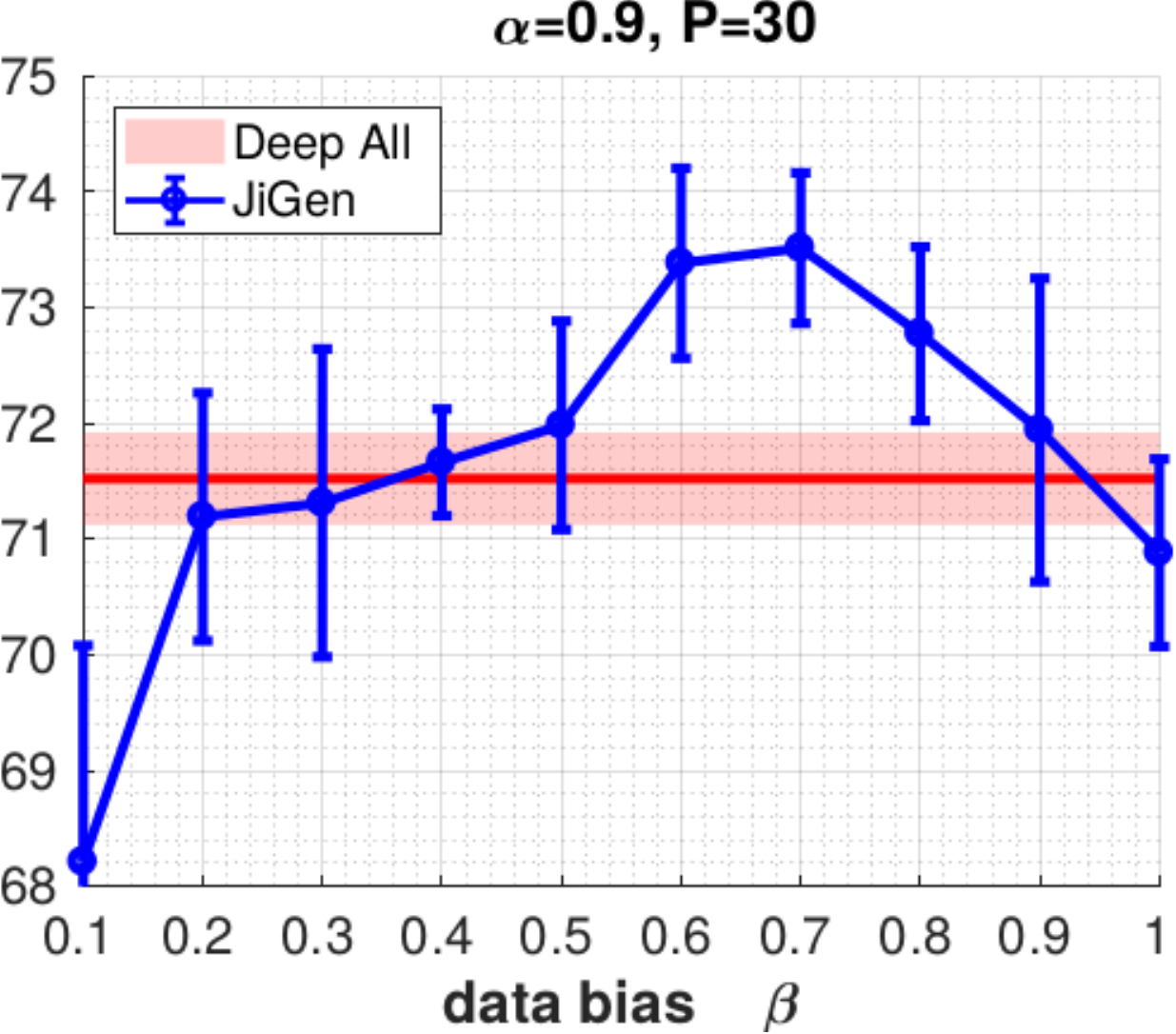} &
\includegraphics[height=3.43cm]{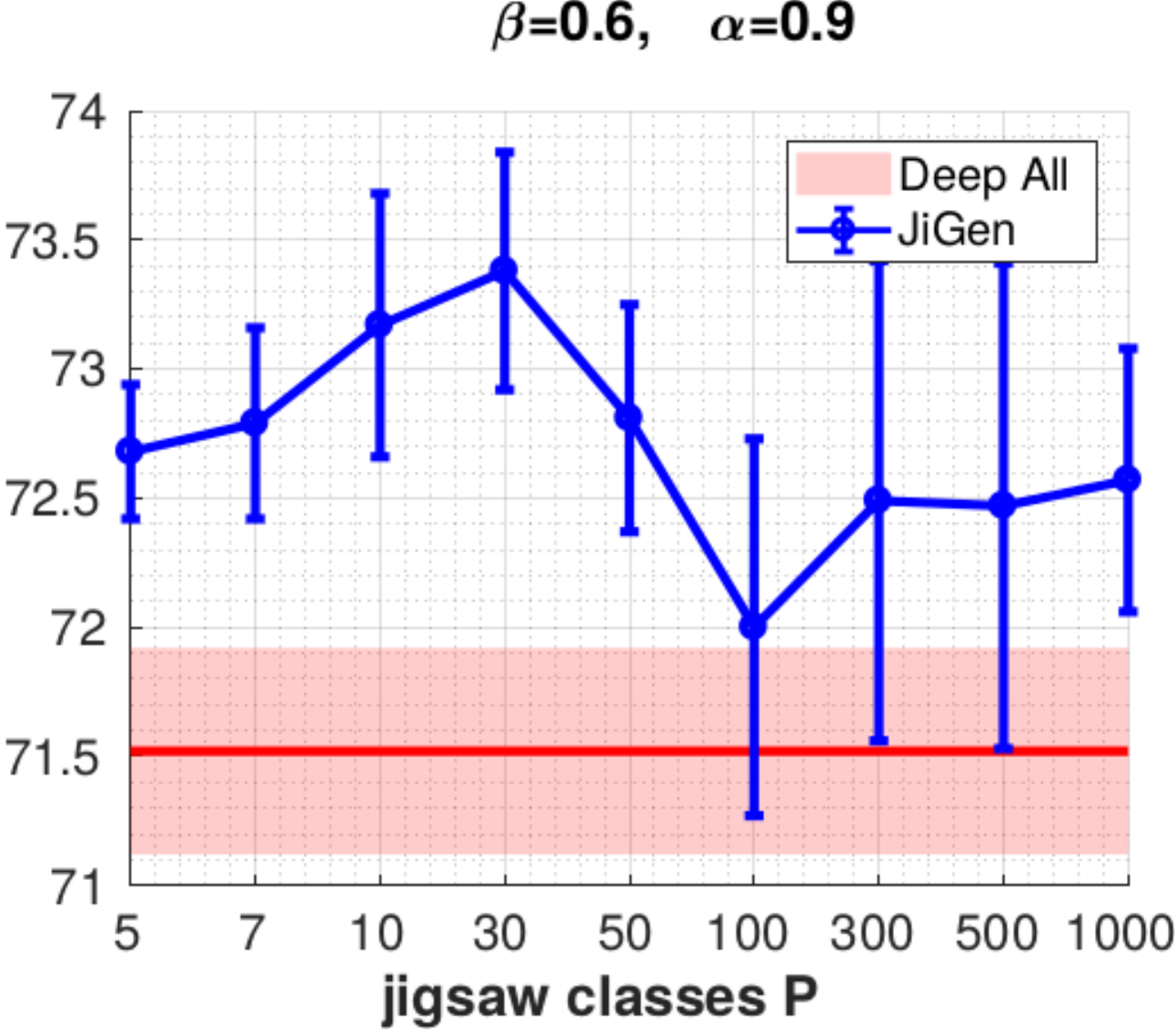}  & \includegraphics[height=3.35cm]{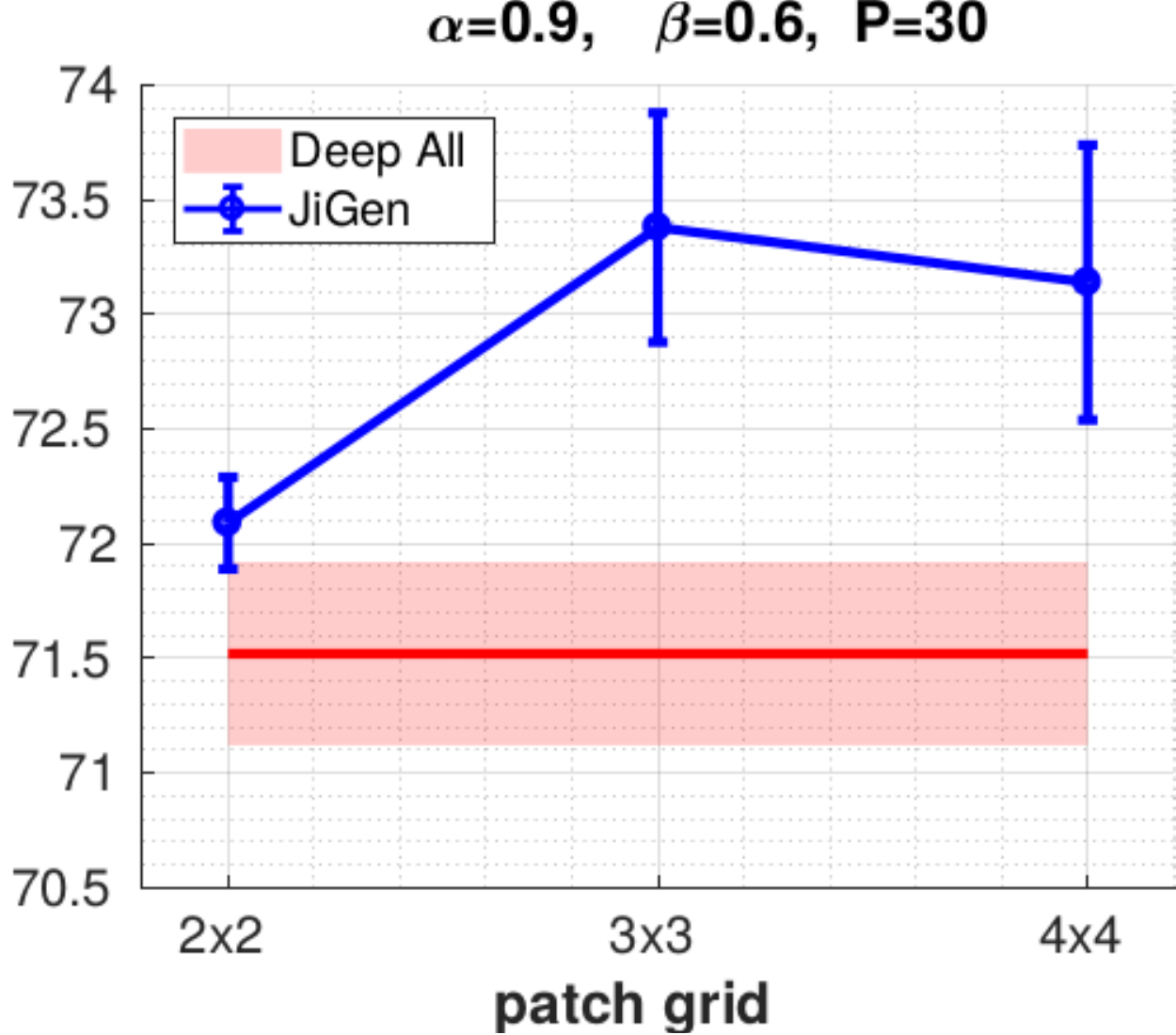}\\
    \end{tabular}
    \caption{Ablation results on the Alexnet-PACS DG setting. The reported accuracy is the global average over all the target domains with three repetitions for each run. The red line represents our \emph{Deep All} average from Table \ref{table:resultsDG_PACS}.}
    \label{fig:ablation}\vspace{-3mm}
\end{figure*}

The central and bottom parts of Table \ref{table:resultsDG_PACS} show the results of \our on the dataset PACS when using 
as backbone architecture Alexnet and Resnet-18\footnote{With Resnet18, to put \our on equal
footing with D-SAM we follow the same data augmentation protocol in \cite{Antonio_GCPR18} and enabled color jittering.}. 
On average \our produces the best result when using Alexnet and it is just slightly worse than the D-SAM reference for Resnet-18. 
Note however, that in this last case, \our outperforms D-SAM in three out of four target cases and the average advantage of 
D-SAM originate only from its result on sketches. 
On average, \our outperforms also the competing methods on the VLCS and on the Office-Home datasets 
(see respectively Table \ref{table:resultsDG_VLCS} and \ref{table:resultsDG_officehome}). 
In particular we remark that VLCS is a tough setting where the most recent works have only presented small gain 
in accuracy with respect to the corresponding Deep All baseline (\eg TF). Since  \cite{Antonio_GCPR18} did not 
present the results of D-SAM on the VLCS dataset, we used the code provided by the authors 
to run these experiments. 
The obtained results show that, although generally able to close large domain gaps across images of different 
styles as in PACS and Office-Home, when dealing with domains all coming from real-world images, the use of 
aggregative modules does not support generalization.

\begin{figure}[tb]\hspace{-5mm}
    \begin{tabular}{c@{~}c}
\includegraphics[height=3cm]{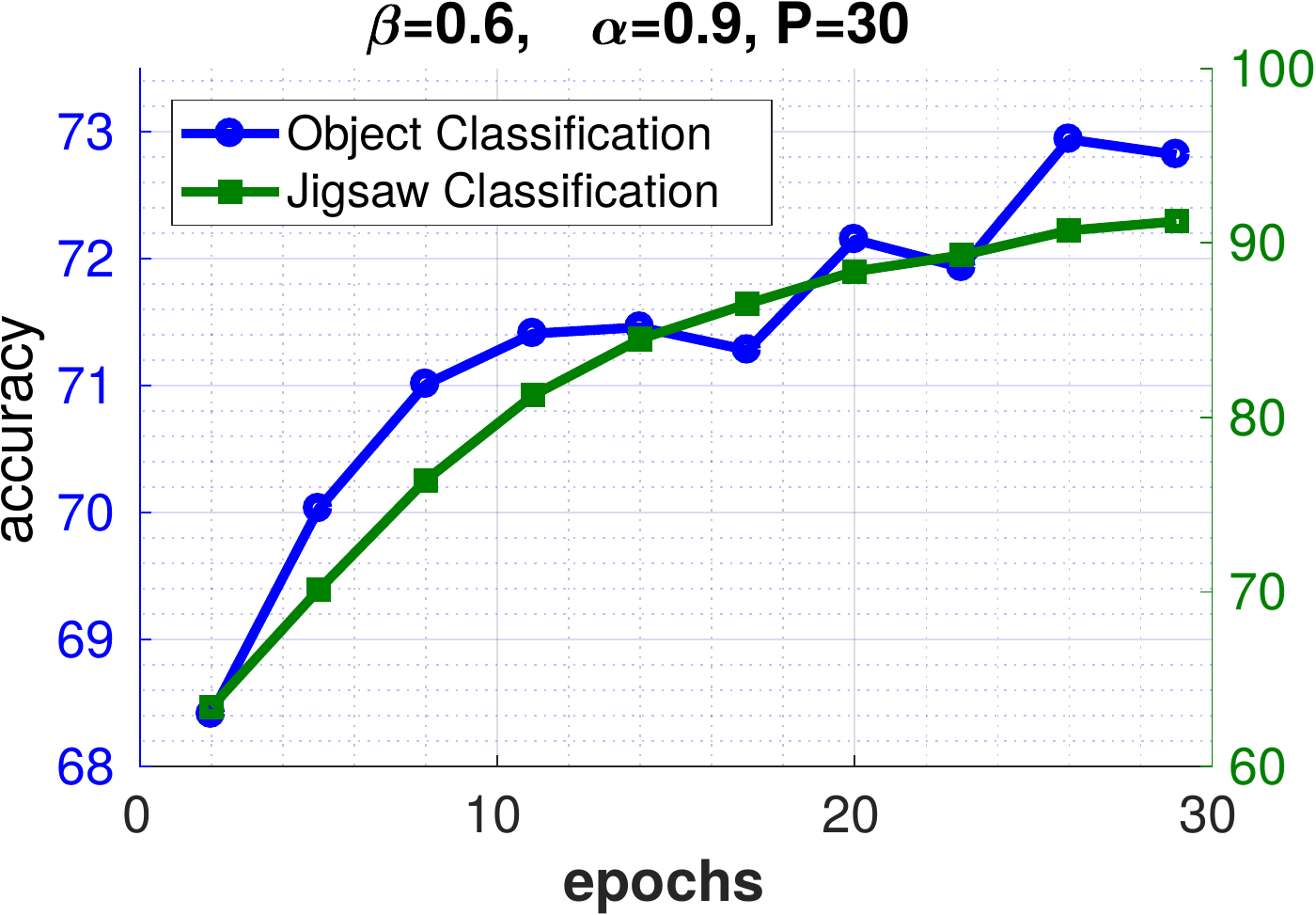}  &  \includegraphics[height=3cm]{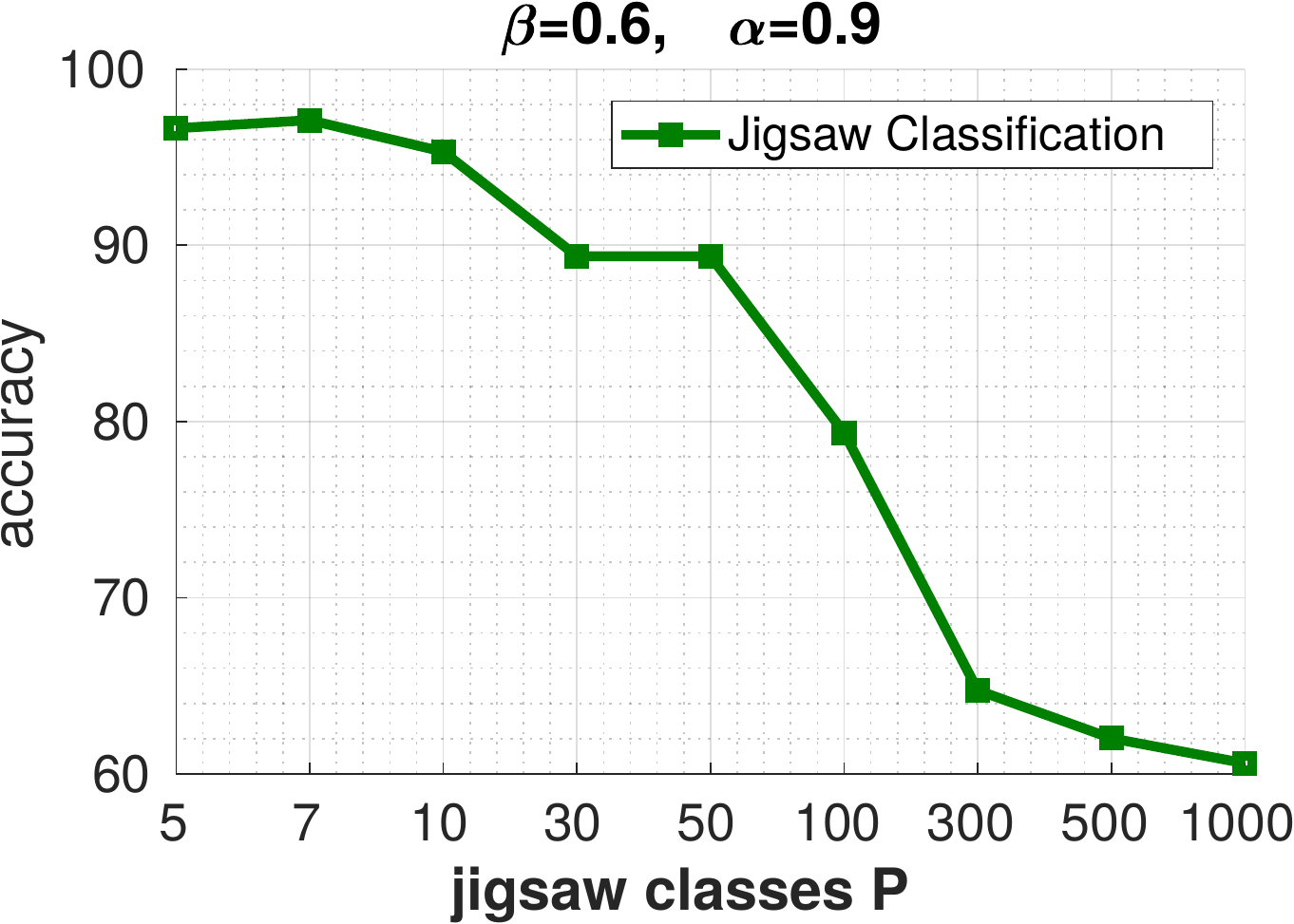}\\
    \end{tabular}
    \caption{Analysis of the behaviour of the jigsaw classifier on the Alexnet-PACS DG setting. For the plot on the left each axes refers to the color matching curve in the graph.}
    \label{fig:ablation_jigsaw}\vspace{-3mm}
\end{figure}

\vspace{-4mm} \paragraph{Ablation}
We focus on the Alexnet-PACS DG setting for an ablation analysis on the respective roles 
of the jigsaw and of the object classification task in the learning model. For these experiments we kept the 
jigsaw hyperparameters fixed with a $3 \times 3$ patch grid and $P=30$ jigsaw classes. 
$\{\alpha=0, \beta=1\}$ means that the jigsaw task is off, and the data batches contain only original ordered images, which corresponds to \emph{Deep All}.
The value assigned to the data bias $\beta$ drives the overall training: it moves the focus from jigsaw when using low
values ($\beta<0.5$) to object classification when using high values ($\beta\geq0.5$).
By setting the data bias to $\beta=0.6$ we feed the network with more ordered than shuffled images, thus keeping the classification 
as the primary goal of the network. In this case, when changing the jigsaw loss weight $\alpha$ in $\{0.1,1\}$, we observe results
which are always either statistically equal or better than the Deep All baseline as shown in the first plot of Figure \ref{fig:ablation}.
The second plot indicates that, for high values of $\alpha$, tuning  $\beta$ has a significant effect on the overall performance.
Indeed $\{\alpha\sim1, \beta=1\}$ means that jigsaw task is on and highly relevant in the learning process, but we are
feeding the network only with ordered images: in this case the jigsaw task is trivial and forces the 
network to recognize always the same permutation class which, instead of regularizing the learning process,
may increase the risk of data memorization and overfitting. Further experiments confirm that, for $\beta=1$ but lower
$\alpha$ values, \our and Deep All perform equally well. 
Setting $\beta=0$ means feeding the network only with shuffled images. For each image we have $P$ variants, 
only one of which has the patches in the correct order and is allowed to enter the object classifier, resulting
in a drastic reduction of the real batch size. In this condition the object classifier is unable to converge, 
regardless of whether the jigsaw classifier is active ($\alpha>0$) or not ($\alpha=0$). In those cases the accuracy 
is very low ($<20\%$), so we do not show it in the plots to ease the visualization.

\vspace{-4mm} \paragraph{Jigsaw hyperparameter tuning}
By using the same experimental setting of the previous paragraph, the third plot in Figure \ref{fig:ablation} shows the change in performance when the number of jigsaw
classes $P$ varies between 5 and 1000. We started from a low number, with the same order of 
magnitude of the number of object classes in PACS, and we grew till 1000 which is the number used
for the experiments in \cite{NorooziF16}. We observe an overall variation of 1.5 percentage points
in the accuracy which still remains (almost always) higher than the Deep All baseline. 
Finally, 
we ran a test to check the accuracy when changing the grid size and consequently the patch number. Even in this case, the range of variation is limited when passing from a $2\times2$ to a $4\times4$ grid, confirming the conclusions of robustness already obtained for this parameter in \cite{NorooziF16} and \cite{Cruz2017}. Moreover all the results are better than the Deep All reference.

It is also interesting to check whether the jigsaw classifier is producing meaningful results per-se, besides 
supporting generalization for the object classifier. We show its recognition accuracy when 
testing on the same images used to evaluate the object classifier but with shuffled patches. In Figure
\ref{fig:ablation_jigsaw}, the first plot shows the accuracy over the learning epochs
for the object and jigsaw classifiers indicating that both grows simultaneously (on different scales).
The second plot shows the jigsaw recognition accuracy when changing the number of permutation classes $P$: 
of course the performance decreases when the task becomes more difficult, but overall
the obtained results indicate that the jigsaw model is always effective in reordering the shuffled patches.

\begin{figure*}
\centering
\includegraphics[width=0.246\textwidth]{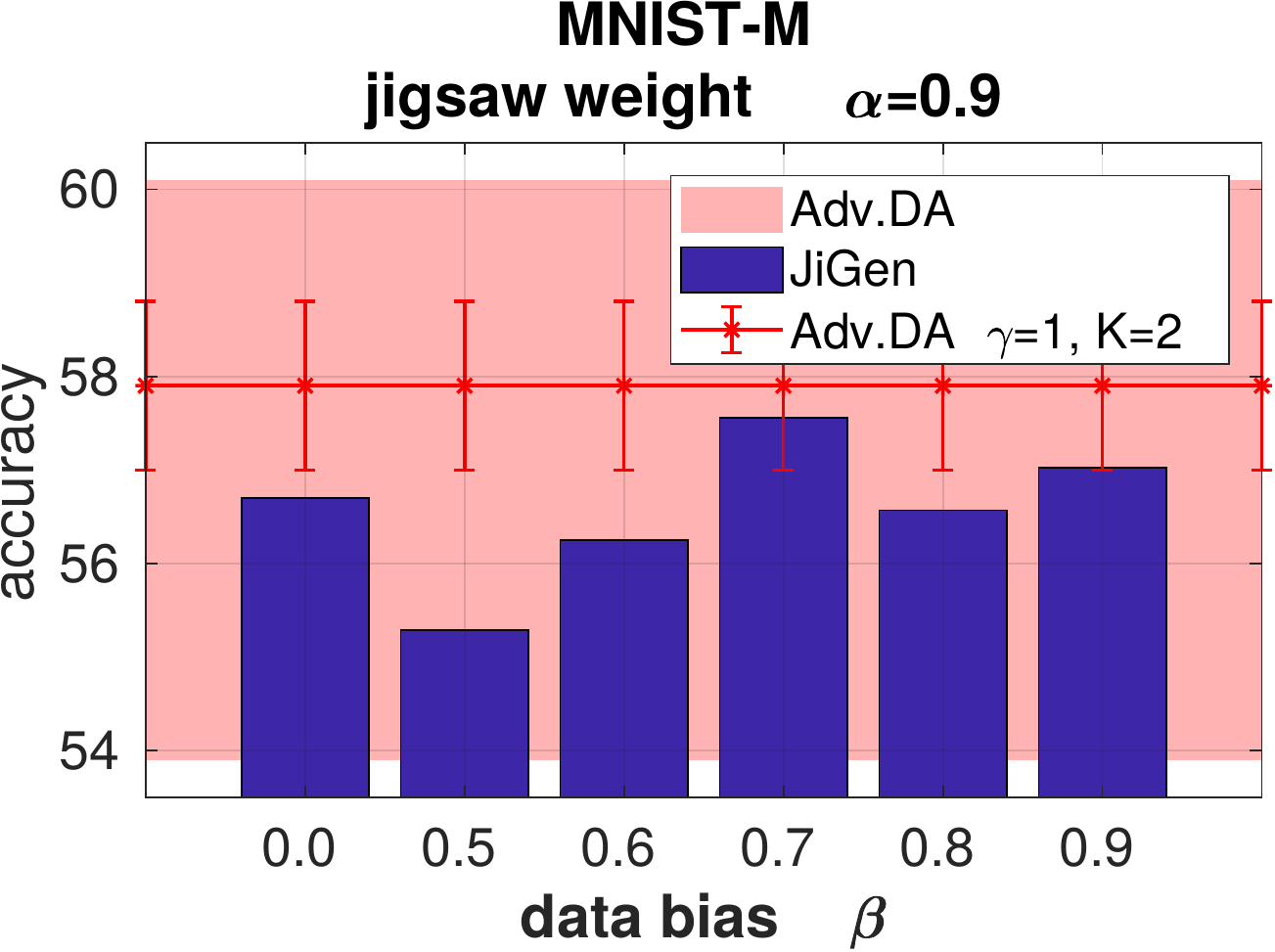}
\includegraphics[width=0.246\textwidth]{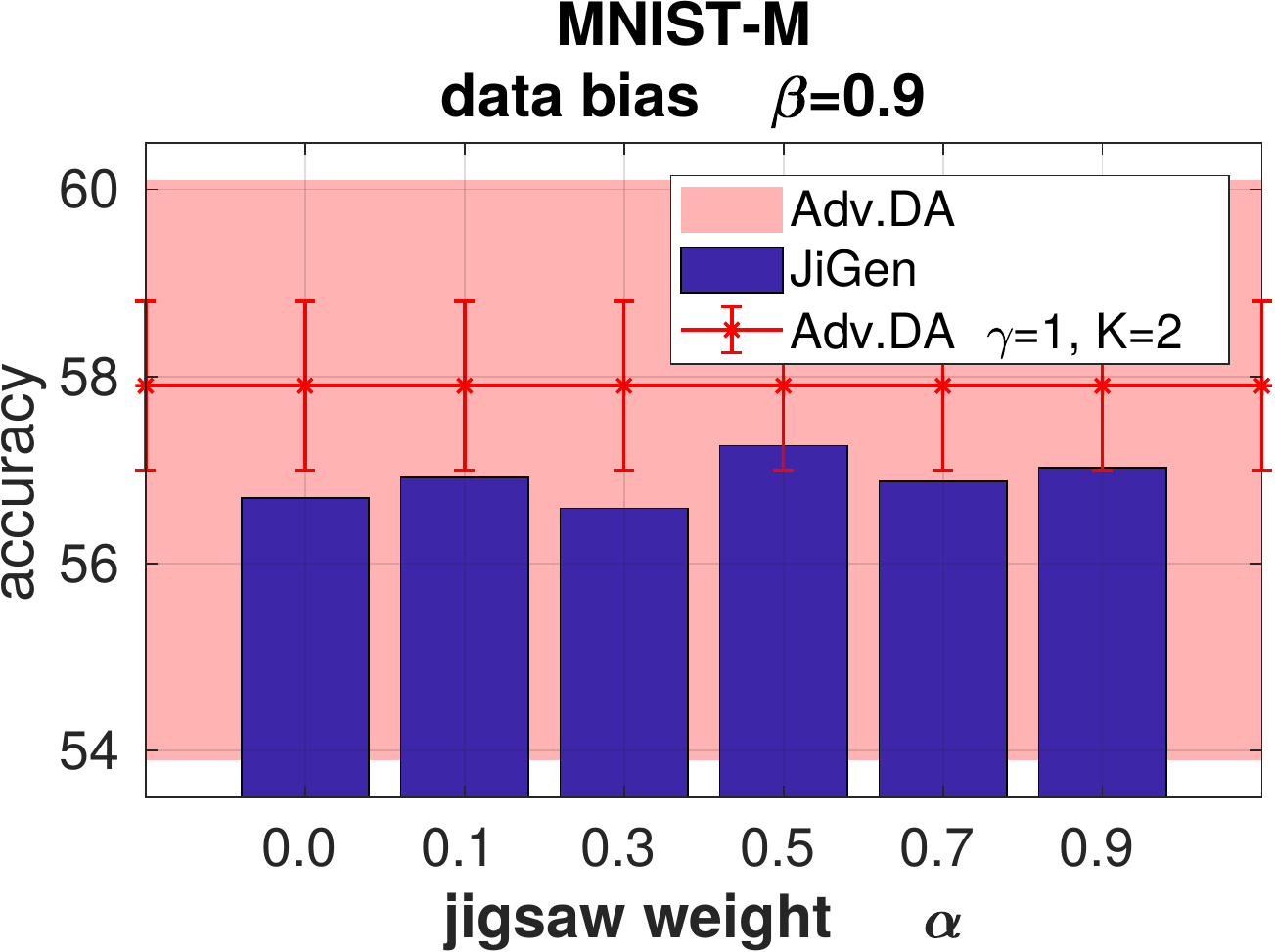}
\includegraphics[width=0.246\textwidth]{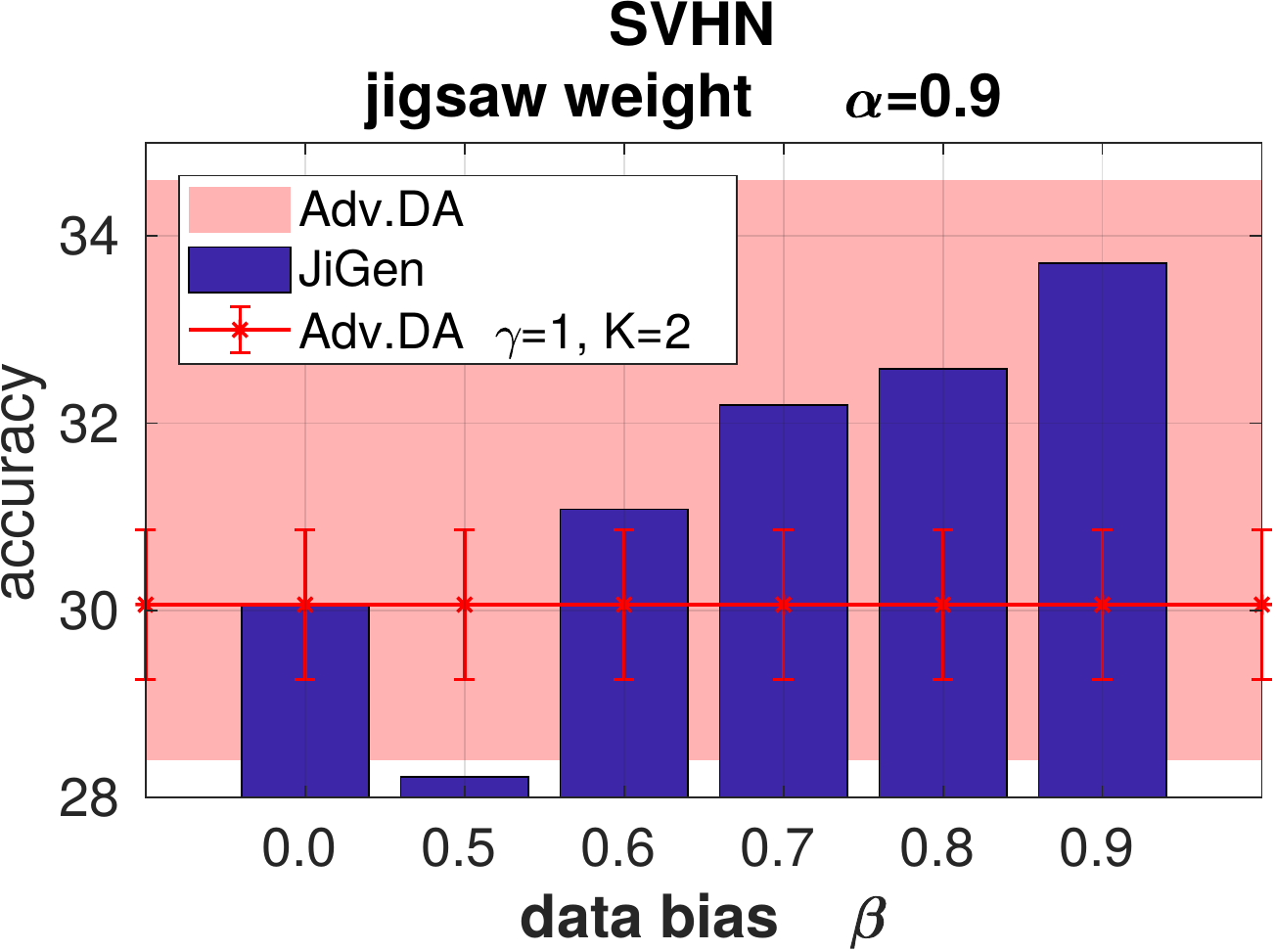}  
\includegraphics[width=0.246\textwidth]{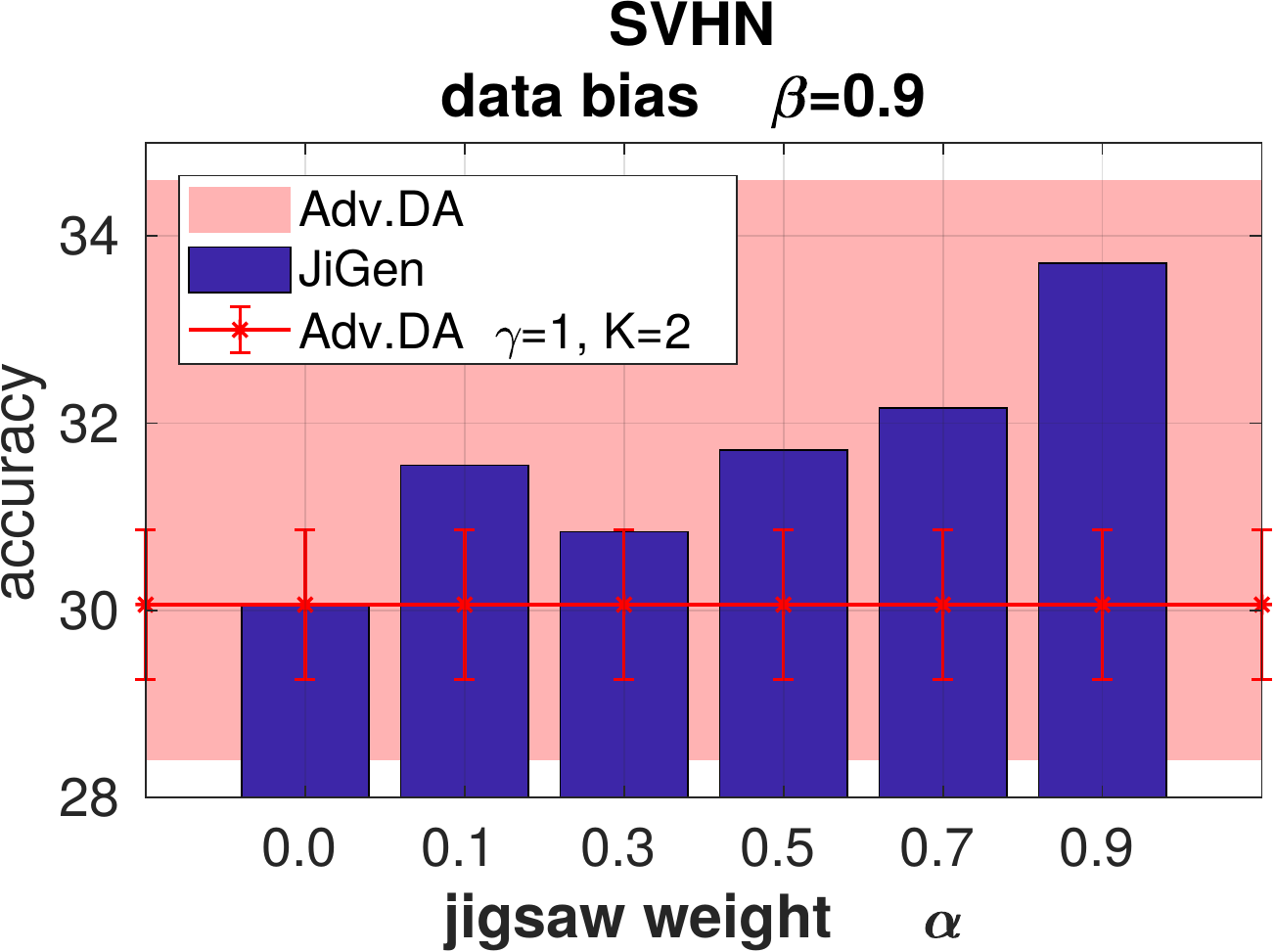}
    \caption{Single Source Domain Generalization experiments. We analyze the performance
    of \our in comparison with the method Adv.DA \cite{Volpi_2018_NIPS}. The
    shaded background area covers the overall range of results of Adv.DA obtained when changing
    the hyper-parameters of the method. The reference result of Adv.DA ($\gamma=1$, $K=2$) together 
    with its standard deviation is indicated here by the horizontal red line.
    The blue histogram bars show the performance of \our when changing the jigsaw weight $\alpha$
    and data bias $\beta$.}
    \label{fig:singesource}\vspace{-3mm}
\end{figure*}

\vspace{-4mm}\paragraph{Single Source Domain Generalization}
The generalization ability of a model depends both on the chosen learning process and on the used training data. 
To investigate the former and better evaluate the regularization effect provided by
the jigsaw task, we  consider the case of training data from a single source domain.
For these experiments we compare against the generalization method based on adversarial data augmentation (\textbf{Adv.DA}) recently presented in  \cite{Volpi_2018_NIPS}. This work proposes an iterative procedure
that perturbs the samples to make them hard to recognize under the current model and then combine them with the original ones while solving the classification task.
We reproduced the experimental setting used  in \cite{Volpi_2018_NIPS} and adopt a similar 
result display style with bar plots for experiments on the MNIST-M and SVHN target datasets when training on MNIST.
In Figure \ref{fig:singesource} we show the performance of \our when varying the data bias $\beta$ 
and the jigsaw weight $\alpha$. With the red background shadow we indicate the overall range 
covered by Adv.DA results when changing its parameters\footnote{The whole set of results is provided as supplementary material of \cite{Volpi_2018_NIPS}.}, while the horizontal line is the reference Adv.DA results
around which the authors of \cite{Volpi_2018_NIPS} ran their parameter ablation analysis. The figure indicates that,
although Adv.DA can reach high peak values, it is also very sensitive to the chosen hyperparameters. 
On the other hand, \our is much more stable and it is always better than the lower accuracy value of Adv.DA with 
a single exception for SVHN and data bias 0.5, but we know from the ablation analysis, that this
corresponds to a limit case for the proper combination of object and jigsaw classification.
Moreover, \our gets close to Adv.DA reference results for MNIST-M and significantly outperform it for SVHN.

\begin{table}[tb]
\begin{center} \small
\begin{tabular}{@{}c@{}c@{}c@{~~}c@{~~}c@{~~}c|@{~~}c}
\hline
\multicolumn{2}{@{}c@{}}{\textbf{PACS-DA}}  & \textbf{art\_paint.} & \textbf{cartoon} &  \textbf{sketches} & \textbf{photo} &   \textbf{Avg.}\\ \hline
\multicolumn{7}{@{}c@{}}{\textbf{Resnet-18}}\\
\hline
\multirow{3}{*}{\cite{mancini2018boosting}}& Deep All & 74.70 & 72.40 & 60.10 & 92.90 & 75.03\\
& Dial & 87.30 & 85.50 & 66.80 & 97.00 &  84.15 \\
& DDiscovery  &  \textbf{87.70}	& \textbf{86.90}	& 69.60 & 97.00		& 85.30\\
\hline
& Deep All & 77.85  & 74.86  & 67.74  & 95.73  & 79.05 \\
& \hspace{-3mm}\textbf{\our$_{\alpha^s=\alpha^t=0.7}$} &  84.88 &	81.07 &	\textbf{79.05} &	\textbf{97.96} &	\textbf{85.74} \\
\hline
& \our$_{\alpha^t=0.1}$ &  85.58 &	82.18 &	78.61  &	98.26 &	86.15 \\
& \our$_{\alpha^t=0.3}$ &  85.08 &	81.28 &	81.50 &	97.96 &	86.46 \\
& \our$_{\alpha^t=0.5}$ &  85.73 &	82.58 &	78.34 &	98.10 &	86.19 \\
& \our$_{\alpha^t=0.9}$ &  85.32 &	80.56 &	79.93 &	97.63 &	85.86 \\
\hline
\end{tabular}
\caption{Multi-source Domain Adaptation results on PACS obtained as average over three repetitions for each run. Besides considering the same jigsaw loss weight for source and target samples $\alpha^s=\alpha^t$, we also tuned the target jigsaw loss weight while keeping $\alpha^s=0.7$ showing that we can get even higher results.}
\label{table:resultsDA_PACS}
\end{center}\vspace{-8mm}
\end{table}

\vspace{-4mm}\paragraph{Unsupervised Domain Adaptation}
When unlabeled target samples are available at training time we can let the jigsaw puzzle task involve these data.
Indeed patch reordering does not need image labels and running the jigsaw optimization process on both source 
and target data may positively influence the source classification model for adaptation. 
To verify this intuition we considered again the PACS dataset and used it in the same unsupervised domain 
adaptation setting of \cite{mancini2018boosting}. This previous work proposed a method to first discover 
the existence of multiple latent domains in the source data and then differently adapt their knowledge to the
target depending on their respective similarity. It has been shown that this domain discovery 
(\textbf{DDiscovery}) technique outperforms other powerful adaptive approaches as \textbf{Dial} 
\cite{carlucci2017just} when the source actually includes multiple domains. Both these methods exploit 
the minimization of the entropy loss as an extra domain alignment condition: in this way
the source model when predicting on the target samples is encouraged to assign maximum prediction 
probability to a single label rather than distributing it over multiple class options. 
For a fair comparison we also turned on the entropy loss for \our with weight $\eta=0.1$.
Moreover, we considered two cases for the jigsaw loss: either keeping the weight $\alpha$ already
used for the PACS-Resnet-18 DG experiments for both the source and target data ($\alpha=\alpha^s=\alpha^t=0.7$),
or treating the domain separately with a dedicated weight for the jigsaw target loss
($\alpha^s=0.7$, $\alpha^t=[0.1,0.3,0.5,0.9]$). 
The results for this setting are summarized in Table \ref{table:resultsDA_PACS}. The obtained accuracy 
indicates that \our outperforms the competing methods on average and in particular on the difficult task of recognizing sketches. Furthermore, the advantage remains
true regardless of the specific choice of the target jigsaw loss weight.

\section{Conclusions}
In this paper we showed for the first time that generalization across visual domains can be achieved effectively by learning to classify and learning intrinsic image invariances at the same time. We focused on learning spatial co-location of image parts, and proposed a simple yet powerful framework that can accommodate a wide spectrum of pre-trained convolutional architectures. Our method \our can be 
seamlessly and effectively used for domain adaptation and generalization as shown by the experimental results.

We see this paper as opening the door to a new research thread in domain adaptation and generalization. While here we focused on a specific type of invariance, several other regularities could be learned possibly leading to an even stronger benefit. Also, the simplicity of our approach calls for testing its effectiveness in applications different from object categorization, like semantic segmentation and person re-identification, where the domain shift effect strongly impact the deployment of methods in the wild.  

\vspace{-4mm}\paragraph{Acknowledgments}
This work was supported by the ERC grant 637076 RoboExNovo and a NVIDIA Academic Hardware Grant.

\appendix
\section{Appendix}

\emph{We provide here some further analysis and experimental results on using jigsaw puzzle and other self-supervised tasks as auxiliary objectives to improve generalization across visual domains.}

\begin{figure}
\begin{tabular}{c@{~}c@{~}c@{~}c@{~}c@{~}c}
& \footnotesize{Deep All \ding{55}} & \footnotesize{JiGen \ding{51}}&  & \footnotesize{Deep All  \ding{55}} & \footnotesize{JiGen \ding{51}}\\
\includegraphics[width=0.145\linewidth,frame]{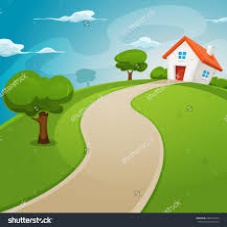} &
\includegraphics[width=0.145\linewidth,frame]{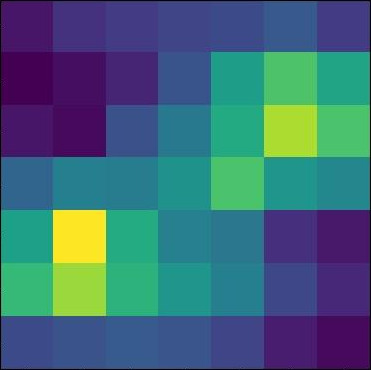} &
\includegraphics[width=0.145\linewidth,frame]{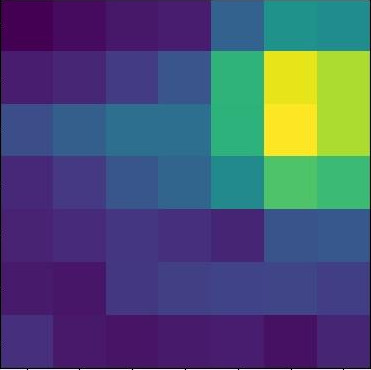} &
\includegraphics[width=0.145\linewidth,frame]{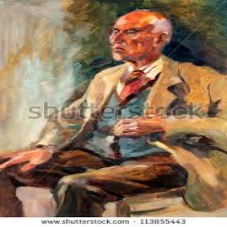} &
\includegraphics[width=0.145\linewidth,frame]{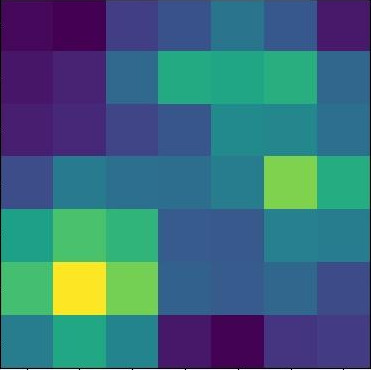}&
\includegraphics[width=0.145\linewidth,frame]{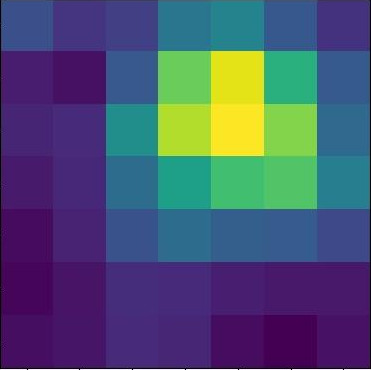}\\ 
\includegraphics[width=0.145\linewidth,frame]{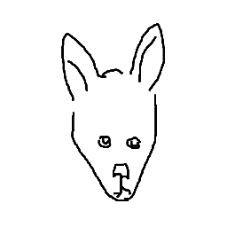} &
\includegraphics[width=0.145\linewidth,frame]{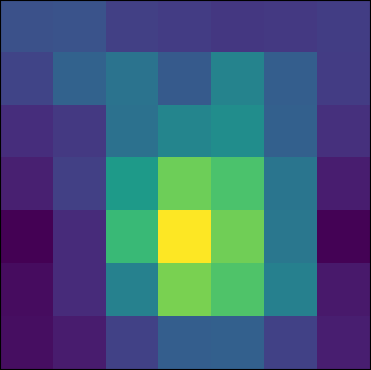} &
\includegraphics[width=0.145\linewidth,frame]{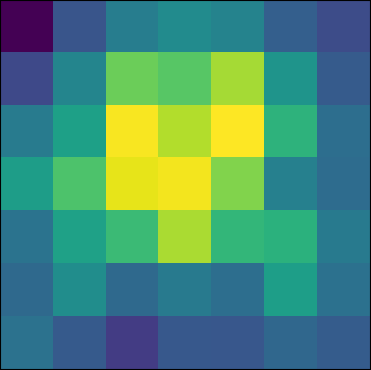} &
\includegraphics[width=0.145\linewidth,frame]{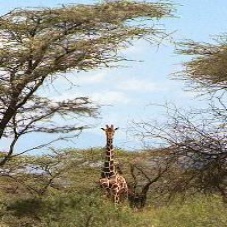} &
\includegraphics[width=0.145\linewidth,frame]{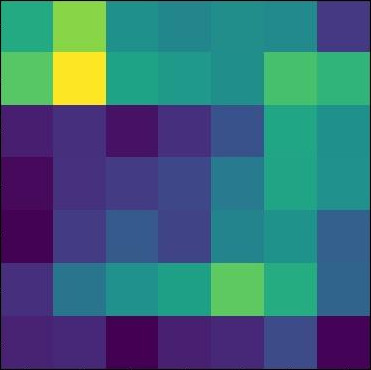}&
\includegraphics[width=0.145\linewidth,frame]{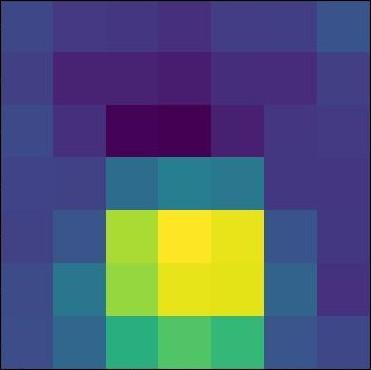}\vspace{-1mm}\\ 
& \footnotesize{Deep All \ding{51}} & \footnotesize{JiGen \ding{55}}&  & \footnotesize{Deep All  \ding{51}} & \footnotesize{JiGen \ding{55}}\\
\includegraphics[width=0.145\linewidth,frame]{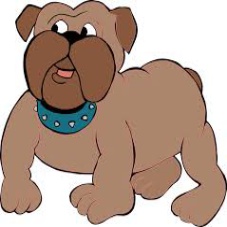} &
\includegraphics[width=0.145\linewidth,frame]{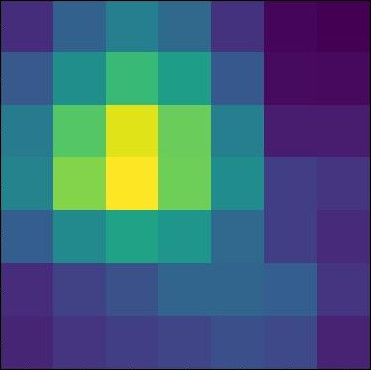} &
\includegraphics[width=0.145\linewidth,frame]{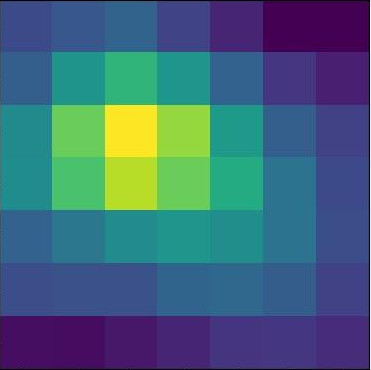} &
\includegraphics[width=0.145\linewidth,frame]{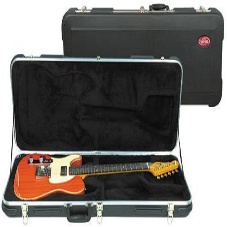} &
\includegraphics[width=0.145\linewidth,frame]{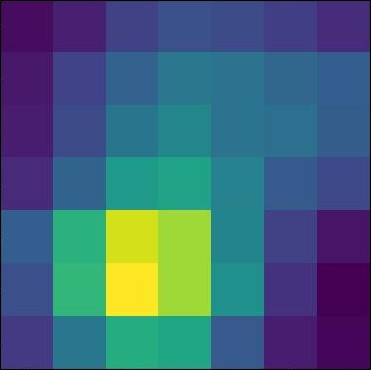}&
\includegraphics[width=0.145\linewidth,frame]{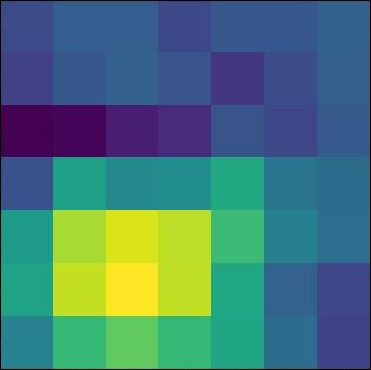}\
\end{tabular}\caption{CAM activation maps: yellow corresponds to high values, while dark blue corresponds to low values. \our is able to localize the most informative part of the image, useful for object class prediction regardless of the visual domain.}\label{fig:examples}
\end{figure}

\vspace{-4mm}\paragraph{Visual explanation and Failure cases }
The relative position of each image patch with respect to the others captures visual regularities which are at the same time shared among domains and discriminative with respect to the object classes.
Thus, by solving jigsaw puzzles we encourage the network to localize and re-join relevant object sub-parts regardless of the visual domain. This helps to focus on the most informative image areas.
For an in-depth analysis of the learned model we adopted the Class Activation Mapping (CAM, \cite{zhou2016cvpr}) method on ResNet-18, with which we produced the activation maps in Figure \ref{fig:examples} for the PACS dataset.
The first two rows show that JiGen is better at localizing the object class with respect to Deep All. The last row indicates that the mistakes are related to some flaw in data interpretation, while the localization remains correct. 

\vspace{-4mm}\paragraph{Self-supervision by predicting image rotations} 
Reordering image patches to solve jigsaw puzzle is not the only self-supervised approach that can be combined with supervised learning for domain generalization. We ran experiments by using as auxiliary self-supervised task the rotation classifier (four classes $[0\degree, 90\degree, 180\degree, 270\degree]$) proposed in \cite{gidaris2018unsupervised}. We focused on the PACS dataset with the Alexnet-based architecture,  
following the same  protocol used for JiGen.
The obtained accuracy (Table \ref{table:resultsROT}) is higher than the Deep All baseline, but still lower than what obtained with our method. Indeed object 2d orientation provides useful semantic information when dealing with real photos, but it becomes less critical for cartoons and sketches.

\begin{table}[tb]
\begin{center} \small
\begin{tabular}{@{}c@{~~~}c@{~~~}c@{~~~}c@{~~~}c@{~~~}c|@{~~~}c}
\hline
\multicolumn{2}{c}{\textbf{PACS}}  & \textbf{art\_paint.} & \textbf{cartoon} &  \textbf{sketches} & \textbf{photo} &   \textbf{Avg.}\\ \hline
\multicolumn{7}{@{}c@{}}{\textbf{Alexnet} }\\
\hline
& Deep All & 66.68 & 69.41 & 60.02 & \underline{89.98}  & 71.52\\
& Rotation & \textbf{67.67}  & 69.83  & 61.04  &  \textbf{89.98}  & 72.13\\
& JiGen & 67.63 & \textbf{71.71} & \textbf{65.18} & 89.00 & \textbf{73.38}\\
\hline
\end{tabular}
\begin{tabular}{c@{~~}c@{~~}c@{~~}c@{~~}c@{~~}c}
\hspace{-1mm}\includegraphics[width=0.145\linewidth,frame]{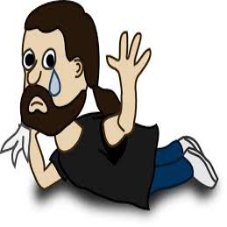} &
\includegraphics[width=0.145\linewidth,frame]{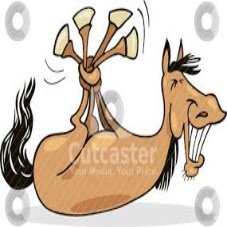} &
\includegraphics[width=0.145\linewidth,frame]{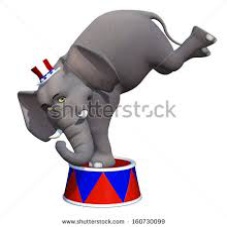} &
\includegraphics[width=0.145\linewidth,frame]{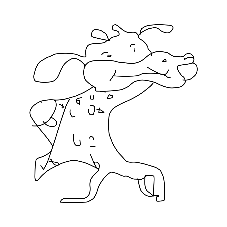} &
\includegraphics[width=0.145\linewidth,frame]{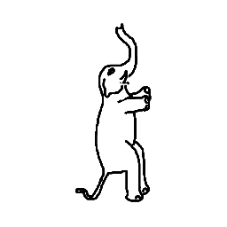}&
\includegraphics[width=0.145\linewidth,frame]{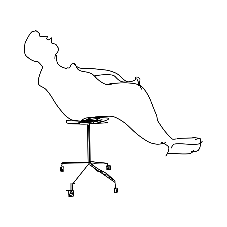}\\ \hline
\end{tabular}
\caption{\emph{Top}: results obtained by using Rotation recognition as auxiliary self-supervised task. \emph{Bottom}: three cartoons and three sketches that show objects with odd orientations.}
\label{table:resultsROT}
\end{center}
\end{table}

{\small
\bibliographystyle{ieee_fullname}
\bibliography{egbib}
}

\end{document}